\documentclass[11pt, logo, onecolumn, copyright, colorlinks=true, allcolors=blue]{nvidiatechreport}

\usepackage{amsmath,amsfonts,bm}

\def\eqref#1{equation~\ref{#1}}

\def\1{\bm{1}}

\DeclareMathAlphabet{\mathsfit}{\encodingdefault}{\sfdefault}{m}{sl}
\SetMathAlphabet{\mathsfit}{bold}{\encodingdefault}{\sfdefault}{bx}{n}

\usepackage{url}
\usepackage{hyperref}
\usepackage{graphicx}
\usepackage{subcaption,ragged2e}
\usepackage[round]{natbib}
\usepackage{placeins}
\usepackage{todonotes}

\usepackage[normalem]{ulem}
\usepackage[utf8]{inputenc}
\usepackage[T1]{fontenc}
\usepackage{tikz}
\usepackage{pgf-pie}
\usepackage{enumitem}
\usetikzlibrary{positioning}
\usetikzlibrary{calc}
\usepackage{array}
\usepackage{booktabs}
\usepackage{wrapfig}
\usepackage{caption}
\usepackage{listings}
\usepackage{adjustbox}
\usepackage{footnote}
\usepackage{multirow}
\usepackage{amsmath}
\usepackage{amsfonts}
\usepackage{breakcites}
\usepackage{blindtext}
\usepackage{svg}
\usepackage[most]{tcolorbox}
\usepackage{tabularx}
\usepackage{graphicx}
\usepackage{xcolor}
\usepackage{float}
\usepackage{needspace}
\usepackage{makecell}
\usepackage{colortbl}
\usepackage{stfloats}
\setlist[itemize]{topsep=1pt,itemsep=1pt}
\setlist[enumerate]{topsep=1pt,itemsep=1pt}
\usepackage{xparse}
\RenewDocumentCommand{\paragraph}{s o m}{%
  \par\medskip\noindent\textbf{#3}\quad\ignorespaces
}

\title{One Model to Loop Them All: Elastic Latent Recurrence for Language Models}
\renewcommand{\Affilfont}{\normalfont\normalsize}
\author{Faisal Ladhak\textsuperscript{*}, Shantanu Acharya\textsuperscript{*}, Santiago Akle Serano, Simeng Sun, \\
Boris Ginsburg, Krishna C Puvvada}
\affil{
\texttt{\{fladhak,shantanua\}@nvidia.com}
}
\date{}
\newcommand{\equalcontributiontext}{\footerfont\textsuperscript{*}Equal contribution.}

\begin{document}

\begin{abstract}
\large \textbf{Abstract}

\normalsize
A latent recurrent language model applies a shared block of layers repeatedly, so one checkpoint can trade inference compute for accuracy by changing the loop count. At one loop, however, such models typically fall well short of a non-recurrent model of the same size, a one-loop tax that makes a separate non-recurrent model the better choice whenever latency matters. We present Elastic Latent Recurrence (ELR), a recipe for training looped language models without this tax. At one loop, an ELR model costs the same to serve as a non-recurrent model of the same size and, trained on the same data, reaches similar accuracy. Each additional loop then buys more accuracy for more inference compute. Three choices make this work: a gated state update with a learned step size, loop counts mixed within each optimizer update, and conditioning each pass on its loop index and budget. The first two preserve one-loop accuracy, and the third lowers loss at every additional loop. Mixed depths can be introduced late in pretraining to limit the added training cost, and with zero-initialized conditioning added at the switch, this late schedule reaches lower loss than mixed-depth training from scratch. In MoE models, reusing first-loop expert routes on later loops cuts mixed-depth update time by 11.7\% with no measurable change in accuracy. At inference, keeping only the last loop's attention KV cache holds attention KV memory at its one-loop level while retaining most of the gain from additional loops. On 4B dense and 30B-A3B MoE hybrid models, ELR at one loop is 0.52 points below and 1.04 points above its matched baseline on a ten-task average, and four loops add 7.25 and 4.75 points. Carried through post-training, the recipe yields Koi-4B, a single checkpoint that scores an estimated 18 on the Artificial Analysis Intelligence Index at one loop, matching Qwen3.5-4B, and 29 at four loops.
\end{abstract}

\maketitle

\section{Introduction}
\label{sec:introduction}

Latent recurrent, or looped, language models (LMs) apply a shared block of layers repeatedly, so a single checkpoint can spend more inference compute per token by running more loops~\citep{dehghani2018universal,geiping2025latent,zhu2025scaling}. When the loop count is varied during training, the same weights can be served at several depths, a property we call elastic depth~\citep{geiping2025latent,jeddi2026loopformer}. This is an attractive property for model deployment. The loop count enables simple control of quality vs test-time-compute tradeoff, similar to reasoning-effort setting for chain-of-thought, with the difference that no extra tokens are generated.

Existing elastic-depth models fall far short of a non-recurrent model of equal depth  at one loop. For example, Huginn LM, trained with a mean recurrence of 32, trails its fixed-depth baseline on nearly every task, scoring 34.9 against 46.4 on ARC-Easy even after 800B tokens~\citep{geiping2025latent}. Parcae is below its Transformer baseline at one loop and above it at eight~\citep{prairie2026parcae}. We call this gap the one-loop tax. A model that pays it must spend several loops recovering ground before additional compute buys anything, so a separately trained non-recurrent model remains the better choice for the latency-sensitive tier. A small loop range does not remove the tax on its own. In our ablations, sampling uniformly over one to five loops still leaves the one-loop endpoint behind the baseline (Section~\ref{sec:ablations_one_loop_rank_override}). Avoiding the tax is a different question from whether looping is a compute-efficient way to scale a model~\citep{prairie2026parcae,schwethelm2026recurrence,gao2026loop}. We take the non-recurrent model as baseline, and ask what it costs to make it elastic without paying the one-loop tax.

\begin{figure}[t]
  \centering
  \includegraphics[width=\linewidth]{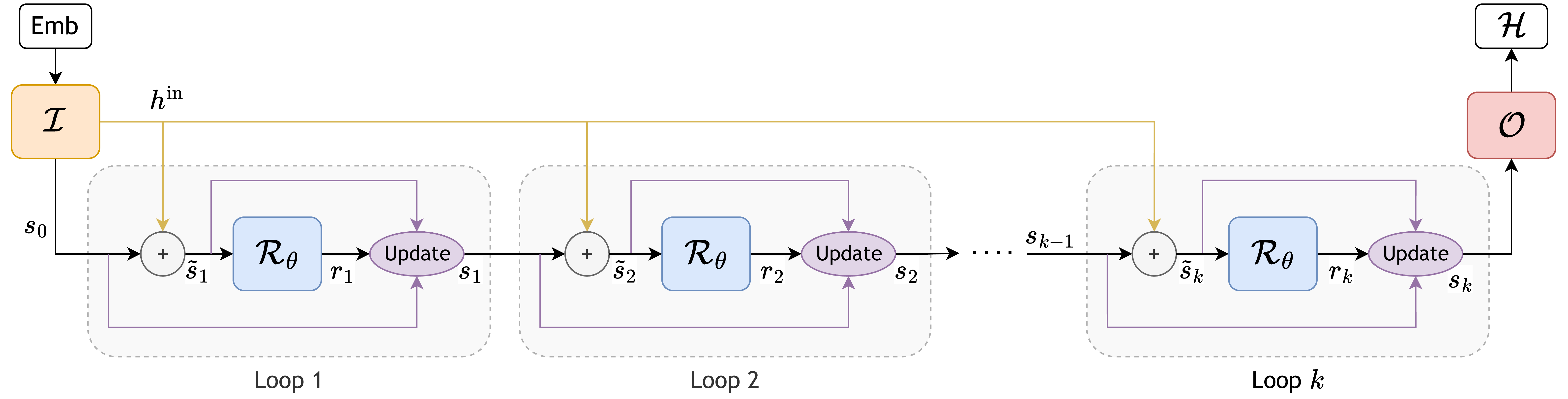}
  \caption{Elastic Latent Recurrence. The shared recurrent block $\mathcal{R}_{\theta}$ is applied $k$ times between the input $\mathcal{I}$ and output $\mathcal{O}$ blocks. Each loop $i$ re-injects $h^{\mathrm{in}}$ and updates the state with the adaptive-delta rule (Eq.~\ref{eq:elr_adaptive_delta_rule}), using a gate computed from $\tilde{s}_i$. At $k=1$, every unique layer executes exactly once.}
  \label{fig:elr_architecture}
\end{figure}

We present \emph{Elastic Latent Recurrence} (ELR), a training recipe for looped language models that avoids the tax. ELR partitions a backbone into input, shared recurrent, and output blocks, so that at one loop every unique layer executes exactly once and the model has the same parameters and inference compute as a non-recurrent model of the same size. Three choices then make the model elastic. The first is the state update. Rather than replacing the state with the recurrent block's output, ELR takes a gated step along the proposed change, with a learned step size per token and channel. The second is where loop counts are mixed. Sampling one loop count per optimizer step and varying it across steps, as in Huginn's locked-step sampling, does not preserve the one-loop endpoint even when one loop is drawn a third of the time. Mixing loop counts within each step does. We implement this by having a rotating subset of data-parallel ranks run one loop on every microbatch.
The first two choices decide whether the one-loop endpoint survives mixed-depth training. The third is conditioning each pass on its loop index and budget, which lets passes specialize and lowers loss at every added loop.
Although recurrent-depth models have been reported to diverge at learning rates that non-recurrent models tolerate~\citep{prairie2026parcae}, every ELR model in this paper trains with its baseline's optimizer settings unchanged.

The recipe is cheap to apply. Mixed-depth training runs the recurrent block several times per example, but that cost need not be paid over the whole run. Training at one loop for the first three quarters of pretraining and at mixed depth only for the last quarter costs an estimated 16\% more training FLOPs than the baseline against 62\% for mixed-depth training throughout (for the 4B configuration (Eq. \ref{eq:delayed_training_cost})). The conditioning heads are zero-initialized at the switch, so the conditioned model starts as exactly the one-loop checkpoint, and with them the late schedule outperforms mixed-depth training from scratch. For recurrent MoE models, \emph{cross-loop route replay} evaluates each router once, on the first loop, and reuses its assignments on later loops, cutting mixed-depth update time by 11.7\% with no measurable change in accuracy. During inference, last-loop KV caching keeps a single cache that each loop overwrites, so attention KV memory stays at the one-loop level at every loop count.

We validate the core of the recipe at two pretraining scales and the full recipe end to end, using a 0.5B model to ablate every design choice. We train ELR alongside non-recurrent baselines on the same data at two scales, a 4B hybrid model for 1T tokens and a 30B-A3B hybrid mixture-of-experts model for 300B tokens. At one loop, ELR is 0.5 points below its baseline on a ten-task average at 4B and 1.0 point above it at 30B-A3B. At four loops it is 6.7 and 5.8 points above. Applying the full recipe end to end yields Koi-4B, a single checkpoint that scores 18 on the Artificial Analysis Intelligence Index at one loop, the level of comparable 4B models, and 29 at four loops, within a fixed 4B-parameter footprint.

Our contributions are as follows:
\begin{itemize}
  \item \textbf{A recipe for elastic depth without a one-loop tax.} A gated state update and within-step mixing of loop counts preserve the one-loop endpoint; ablations show the common alternatives do not. At one loop, ELR reaches similar accuracy to a non-recurrent baseline trained on the same data, and every additional loop of the same checkpoint improves it.
  \item \textbf{Elastic depth acquired late in pretraining.} Mixed-depth training in the final quarter of the run gives the full loop range for an estimated 16\% more training FLOPs than the baseline, and zero-initialized loop conditioning grafted at the switch makes this schedule outperform training at mixed depth from the start.
  \item \textbf{Cross-loop route replay for looped MoE models.} Reusing first-loop expert assignments on later loops cuts mixed-depth update time by 11.7\% in our experiments with no measurable change in accuracy.
  \item \textbf{Koi-4B.} A post-trained elastic-depth model whose single checkpoint spans 18 to 29 on the Artificial Analysis Intelligence Index by loop count alone.
\end{itemize}

\section{Latent Recurrent Models}
\label{sec:background_latent_recurrent}

Latent recurrent models, also called looped models, apply a shared block of layers repeatedly across depth to iteratively refine a hidden state. The idea descends from earlier work on depth-wise weight sharing and adaptive computation~\citep{dehghani2018universal,graves2016adaptive,banino2021pondernet}, and recent language models adopt it to decouple effective depth from unique parameter count~\citep{geiping2025latent,zhu2025scaling,bae2025mixture}.

We describe a latent recurrent model as a composition of three blocks: an input block $\mathcal{I}$ with $L_{\mathrm{in}}$ layers, a recurrent block $\mathcal{R}_{\theta}$ with $L_{\mathrm{rec}}$ layers and parameters $\theta$, and an output block $\mathcal{O}$ with $L_{\mathrm{out}}$ layers. Given a token sequence $x$, the input block produces a fixed representation $h^{\mathrm{in}} = \mathcal{I}(\operatorname{Emb}(x))$, from which a recurrent state is initialized and refined by $k$ applications of the recurrent block:
$$
  s_0 = \operatorname{Init}(h^{\mathrm{in}}), \qquad
  s_i = \mathcal{G}_{\theta}\bigl(s_{i-1}, h^{\mathrm{in}}\bigr), \quad i = 1,\ldots,k
$$
The transition map $\mathcal{G}_{\theta}$ wraps $\mathcal{R}_{\theta}$ and specifies how the state is updated at each iteration, with $\theta$ shared across all $k$ applications. The output block and language-model head $\mathcal{H}$ then map the final state to logits $z = \mathcal{H}(\mathcal{O}(s_k))$. The choices of $\operatorname{Init}$ and $\mathcal{G}_{\theta}$ vary across architectures: Huginn~\citep{geiping2025latent} initializes $s_0$ from a truncated Gaussian and projects the concatenation of $s_{i-1}$ and $h^{\mathrm{in}}$, whereas Ouro~\citep{zhu2025scaling} has no separate input or output blocks, initializing $s_0$ from the token embeddings and repeating the full stack without input re-injection.

The model has $D_{\mathrm{unique}} = L_{\mathrm{in}} + L_{\mathrm{rec}} + L_{\mathrm{out}}$ distinct layers, but a forward pass with loop count $k$ traverses
\begin{equation}
  D_{\mathrm{eff}}(k) = L_{\mathrm{in}} + k\,L_{\mathrm{rec}} + L_{\mathrm{out}}
  \label{eq:effective_depth}
\end{equation}
layers, so each additional loop adds $L_{\mathrm{rec}}$ layers of inference compute and latency at a fixed parameter count. The loop count may be held constant or sampled from a distribution during training. We use the term elastic depth for a model trained over a range of loop counts and served at any $k$ within that range, so that the loop count acts as an inference-time compute knob.
\section{Method}
\label{sec:method}

This section introduces \emph{Elastic Latent Recurrence} (ELR), a latent recurrent architecture and training recipe for serving a single checkpoint at any loop count within its trained range while avoiding the one-loop tax (Figure~\ref{fig:elr_architecture}).

\subsection{Elastic Latent Recurrence}
\label{sec:method_elr}

ELR uses the three-block architecture defined in Section~\ref{sec:background_latent_recurrent}, with approximately the middle half of the layers in $\mathcal{R}_{\theta}$ and the remaining layers split evenly between $\mathcal{I}$ and $\mathcal{O}$. At $k=1$, every unique layer executes exactly once, with $D_{\mathrm{eff}}(1) = D_{\mathrm{unique}}$, matching a non-recurrent model of the same size in parameters and compute.

The recurrent state is initialized as $s_0 = h^{\mathrm{in}}$. At each loop $i=1,\ldots,k$, $\mathcal{G}_{\theta}$ re-injects $h^{\mathrm{in}}$ into the current state, passes the result through $\mathcal{R}_{\theta}$, and applies a state update rule:
$$
  \begin{aligned}
    \tilde{s}_i &= s_{i-1} + h^{\mathrm{in}} \\
    r_i &= \mathcal{R}_{\theta}(\tilde{s}_i) \\
    s_i &= \operatorname{Update}(s_{i-1}, r_i, \tilde{s}_i)
  \end{aligned}
$$

\paragraph{Adaptive-delta update.}
ELR uses an adaptive-delta rule that updates the state as
\begin{equation}
  \begin{aligned}
    s_i &= \operatorname{Update}(s_{i-1}, r_i, \tilde{s}_i) \\
        &= s_{i-1} + \beta_i \odot (r_i - s_{i-1})
  \end{aligned}
  \label{eq:elr_adaptive_delta_rule}
\end{equation}
where the gate $\beta_i$ sets a per-token, per-channel step size along the recurrent block's proposed revision $r_i - s_{i-1}$:
$$
  \beta_i = \sigma(W_{\beta} \tilde{s}_i + b_{\beta}) \odot \operatorname{softplus}(\gamma)
$$
Here $\operatorname{softplus}(\gamma)$ is a learned per-channel base rate, scaled by a sigmoid gate for each token and channel. The parameters $W_{\beta}$, $b_{\beta}$, and $\gamma$ are shared across loops and add $d^2 + 2d$ parameters for hidden dimension $d$, independent of $k$. We initialize $W_{\beta}$ and $b_{\beta}$ to zero and $\gamma$ to 0, giving an initial step size of approximately 0.35. Section~\ref{sec:ablations_state_update_rule} compares adaptive delta with alternative state update rules.

\subsection{Mixed-Depth Training with One-Loop Rank Overrides}
\label{sec:method_mix_training_one_loop_override}

To serve a single checkpoint at multiple loop counts, ELR is trained with mixed depths. For each microbatch, a loop count is sampled uniformly, $p(k) = 1/K$ for $k \in \{1,\ldots,K\}$, and the standard next-token loss is backpropagated through all $k$ applications of the recurrent block. Uniform sampling improves accuracy at $k>1$ but fails to match the non-recurrent baseline at $k=1$ (Section~\ref{sec:ablations_one_loop_rank_override}). We therefore force a rotating fraction $f$ of the data-parallel ranks to use $k=1$ on every microbatch, while the remaining ranks use the sampled count. A rank then executes one loop with marginal probability $p_1$, and the model is trained at a mean loop count $\bar{k}$:
\begin{equation}
  p_1 = \frac{1}{K} + \Bigl(1-\frac{1}{K}\Bigr)f, \qquad
  \bar{k} = f + (1-f)\,\frac{K+1}{2}
  \label{eq:elr_one_loop_marginal}
\end{equation}
Unlike globally raising the sampling probability of $k=1$, the override places one-loop and $k$-loop gradients in the same update whenever the sampled count exceeds one, and data-parallel reduction averages them before the optimizer step (Section~\ref{sec:ablations_sampling}). We choose $f$ for each $K$ so that $p_1$ remains approximately one third, and study sensitivity to this choice in Section~\ref{sec:ablations_one_loop_rank_override}.

\subsection{Loop-Conditioned Modulation}
\label{sec:method_loop_conditioning}

Without explicit loop conditioning, the recurrent block applies the same function at every pass, with no direct signal of its position or the loop budget. We find that adding this signal further improves ELR accuracy at every additional loop. Following prior looped language models \citep{jeddi2026loopformer, chen2026loopmoe}, we condition loop $i$ of $k$ on progress $\tau_i=i/k$ and budget $\kappa=k/K$, where $K$ is the maximum training loop count:
$$
c_i = \mathrm{MLP}\big([\,\phi(\tau_i);\,\phi(\kappa)\,]\big)
$$
Here $\phi$ is a sinusoidal embedding. The output block uses $(\tau,\kappa)=(1,k/K)$, and the input block is unconditioned. Following AdaLN~\citep{peebles2023scalable} and FiLM~\citep{perez2018film}, $c_i$ modulates each recurrent and output layer $f$ in three ways. It scales and shifts the normalized input, adds a low-rank correction with a loop-dependent gate, and rescales the output before the residual addition:
\begin{equation}
  \begin{aligned}
    \tilde{x} &= \big(1+\lambda(c_i)\big)\odot\mathrm{RMSNorm}(x)+\mu(c_i) \\
    \hat{x} &= \tilde{x}+A\big(g(c_i)\odot B\tilde{x}\big) \\
    y &= x+\big(1+\omega(c_i)\big)\odot f(\hat{x})
  \end{aligned}
  \label{eq:loop_conditioned_modulation}
\end{equation}
Here $\lambda$, $\mu$, $\omega$, and $g$ are linear heads of $c_i$, and $B\in\mathbb{R}^{r\times d}$ and $A\in\mathbb{R}^{d\times r}$ are adapter projections, all specific to each layer and shared across loops.

From the second loop onward, entry modulation renormalizes the recurrent input and adds a loop-dependent input re-injection:
$$
  z_i =
  \begin{cases}
    \tilde{s}_i & i=1 \\
    \big(1+\lambda_e(c_i)\big)\odot\mathrm{RMSNorm}(\tilde{s}_i)+\mu_e(c_i)+\eta(c_i)\odot h^{\mathrm{in}} & i\geq2
  \end{cases}
$$
so the first loop receives $z_1=2h^{\mathrm{in}}$, as in the one-loop checkpoint. At exit, a loop-dependent factor scales the adaptive-delta step of $r_i=\mathcal{R}_{\theta}(z_i;c_i)$, whose gate still uses the unmodulated $\tilde{s}_i$:
\begin{align*}
  \beta_i &= \sigma(W_{\beta}\tilde{s}_i+b_{\beta})\odot\operatorname{softplus}(\gamma) \\
  s_i &= s_{i-1}+\big(\pi_i+g_{\mathrm{exit}}(c_i)\big)\odot\beta_i\odot(r_i-s_{i-1})
  \qquad \pi_1=1,\quad\pi_{i>1}=0
\end{align*}
We introduce conditioning at the switch to mixed-depth training (Section~\ref{sec:method_reducing_recurrent_cost}), initializing all conditioning heads to zero. Per-layer modulation is then the identity, the first loop reproduces the checkpoint's update, and later loops leave the state unchanged through zero exit factors. The model therefore initially preserves the one-loop checkpoint's function at every $k$, in exact arithmetic. Since $c_i$ depends only on $(i,k)$, conditioning vectors and head outputs can be precomputed at inference.

\subsection{Reducing the Cost of Recurrence}
\label{sec:method_reducing_recurrent_cost}

\paragraph{Delayed mixed-depth training.}
Mixed-depth training raises the compute of each step to about $D_{\mathrm{eff}}(\bar{k})/D_{\mathrm{eff}}(1)$ times that of a one-loop step (Eqs.~\ref{eq:effective_depth} and~\ref{eq:elr_one_loop_marginal}). To limit this overhead, we train at $k=1$ for most of the token budget and apply the schedule of Section~\ref{sec:method_mix_training_one_loop_override} only in a final continuation phase. The two phases use the same architecture and differ only in the distribution over k, so the adaptive-delta gate is trained from the start, and the optimizer state and learning-rate schedule carry over unchanged. If the mixed-depth phase covers a fraction $\rho$ of the training tokens, total training compute relative to the non-recurrent baseline is approximately
\begin{equation}
  (1-\rho) + \rho\,\frac{D_{\mathrm{eff}}(\bar{k})}{D_{\mathrm{eff}}(1)}
  \label{eq:delayed_training_cost}
\end{equation}
Section~\ref{sec:ablations_delayed_mix_depth_training} studies how $\rho$ trades one-loop quality against gains at additional loops.

\paragraph{Cross-loop route replay.}
In a recurrent MoE model, every loop re-executes the router of each MoE layer in the recurrent block. Cross-loop route replay instead computes each layer's expert assignments and routing weights on the first loop and reuses them on subsequent loops of the same forward pass. The selected experts still process each loop's updated representations, so replay removes repeated router evaluation while retaining expert dispatch, communication, computation, and output combination. During training, the load-balancing auxiliary loss is computed once on the first loop and scaled by the realized loop count $k$, matching the total weight it receives when accumulated over all $k$ loops under dynamic routing. Replay assumes that first-loop routes remain useful at later loops. Section~\ref{sec:experiments_route_replay} tests this assumption by measuring how routing changes across loops and comparing dynamic routing with replay applied at inference only or during training, and reports the resulting training and inference cost.

\paragraph{Last-loop KV caching.}
During inference, latent recurrent models keep one KV cache per loop for each recurrent attention layer. Last-loop KV caching keeps a single cache that each loop overwrites, so during decoding, the loop-$i$ query attends to earlier tokens' final-loop keys and values and to the current token's own loop-$i$ key and value. Attention KV memory then matches the one-loop model at every $k$. Applied at inference to a model trained with per-loop caches, the scheme assumes final-loop keys and values remain useful to earlier-loop queries. Section~\ref{sec:koi_kv_cache} shows that the scheme retains the gains from additional loops at close to one-loop cache memory.
\section{One-Loop Accuracy and Gains from Additional Loops}
\label{sec:experiments}
This section validates the first two choices of the recipe, the adaptive-delta update and within-step mixing with the delayed schedule, against non-recurrent baselines matched in parameters, data, and training horizon. The models here carry no loop conditioning; Section 5 applies the full recipe to a post-trained model, and Section 6.6 isolates the effect of conditioning.
\begin{figure}[t]
  \centering
  \begin{subfigure}[t]{0.4\linewidth}
    \centering
    \includegraphics[width=\linewidth]{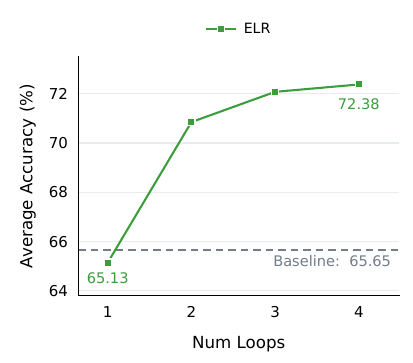}
    \caption{GDP-Hybrid-4B}
    \label{fig:downstream_scaling_gdp_hybrid_4b}
  \end{subfigure}
  \hspace{0.1\linewidth}
  \begin{subfigure}[t]{0.4\linewidth}
    \centering
    \includegraphics[width=\linewidth]{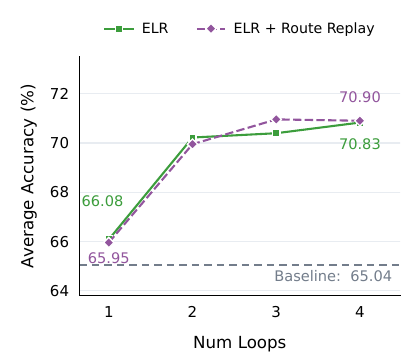}
    \caption{GDP-Hybrid-30B-A3B}
    \label{fig:downstream_scaling_gdp_hybrid_30b_a3b}
  \end{subfigure}
  \caption{Average downstream benchmark accuracy across ten tasks versus inference loop count. Each curve represents a single model evaluated at multiple loop counts. Dashed horizontal lines mark the non-recurrent baselines. Per-task scores are in Tables~\ref{tab:downstream_gdp_hybrid_4b} and~\ref{tab:downstream_gdp_hybrid_30b_a3b}.}
  \label{fig:downstream_scaling}
\end{figure}

\subsection{Experimental Setup}
\label{sec:experiments_setup}

\paragraph{Models.} We evaluate ELR on two hybrid models: GDP-Hybrid-4B, a dense model with approximately 4B parameters, and GDP-Hybrid-30B-A3B, a mixture-of-experts model with approximately 30B total and 3B active parameters. Both share a 52-layer layout of 23 Gated DeltaProduct (GDP)~\citep{siems2025deltaproduct}, 6 self-attention, and 23 feed-forward layers, which are dense MLPs in GDP-Hybrid-4B and MoE layers in GDP-Hybrid-30B-A3B. Each ELR model partitions these layers into 14 input, 24 recurrent, and 14 output layers and is compared with a non-recurrent baseline with the same backbone configuration. All ELR models evaluated in this section follow the architecture of Section~\ref{sec:method_elr} without loop-conditioned modulation. For GDP-Hybrid-30B-A3B, we also train an ELR model with cross-loop route replay (Section~\ref{sec:method_reducing_recurrent_cost}). More details are provided in Appendix~\ref{appendix:architectures}.

\paragraph{Training.} GDP-Hybrid-4B is trained for 1T tokens with AdamW, and GDP-Hybrid-30B-A3B for 300B tokens with Muon and AdamW. Both use a subset of the pretraining data used for Nemotron 3 Ultra~\citep{nvidia2026nemotron3ultra}, with a sequence length of 8,192 and a global batch size of 768. Each non-recurrent baseline is trained on the same data and token budget as its ELR counterpart. All ELR models evaluated in this section use delayed mixed-depth training (Section~\ref{sec:method_reducing_recurrent_cost}), training at one loop for the first 75\% of tokens and then continuing with $k\in\{1,\ldots,4\}$ and a 10\% one-loop rank override, which gives $p_1 = 0.325$ and $\bar{k} = 2.35$ (Eq.~\ref{eq:elr_one_loop_marginal}). By Eq.~\ref{eq:delayed_training_cost}, this schedule has an estimated layer-execution cost of about $1.16\times$ the baseline's. For GDP-Hybrid-30B-A3B, the ELR models with and without cross-loop route replay begin the mixed-depth phase from the same one-loop checkpoint. Full training settings are provided in Appendix~\ref{appendix:model_training_details}.

\paragraph{Evaluation.} Each recurrent curve shows a single final checkpoint evaluated at every loop count in its trained range. We report the downstream accuracy on MMLU, MMLU-Pro, ARC-Challenge, GSM8K, WinoGrande, CommonsenseQA, RACE-high, HumanEval, MBPP, and MATH-500, and report their average. See Appendix~\ref{appendix:evaluation_protocol} for more details.

\subsection{Main Results}
\label{sec:experiments_loop_parity_gains}

\paragraph{One-loop accuracy.} At $k=1$, ELR retains near-baseline accuracy at both scales (Figure~\ref{fig:downstream_scaling}). On GDP-Hybrid-4B, its ten-task average of 65.13 is 0.52 points below the non-recurrent baseline's 65.65. On GDP-Hybrid-30B-A3B, ELR and ELR+Route Replay reach 66.08 and 65.95, slightly above the baseline's 65.04. Per-task scores are reported in Tables~\ref{tab:downstream_gdp_hybrid_4b} and~\ref{tab:downstream_gdp_hybrid_30b_a3b}. At one loop, individual tasks move in both directions; on GDP-Hybrid-4B, ELR trails the baseline by 5.3 points on GSM8K and leads by 5.2 on MATH-500.

\paragraph{Gains from additional loops.} Each additional loop improves the ten-task average accuracy of ELR (Figure~\ref{fig:downstream_scaling}). From $k=1$ to $k=4$, the average rises by 7.25 points for GDP-Hybrid-4B and by 4.75 points for GDP-Hybrid-30B-A3B. The largest per-task gains are on GSM8K (+11.8 points) and MATH-500 (+11.3) for GDP-Hybrid-4B, and on MMLU-Pro (+7.5) and GSM8K (+6.5) for GDP-Hybrid-30B-A3B.

\paragraph{Diminishing returns.} Most of the aggregate gain arrives in the first additional loop. For GDP-Hybrid-4B, two and three loops capture 79\% and 96\% of the gain from $k=1$ to $k=4$. For GDP-Hybrid-30B-A3B, they capture 87\% and 91\%. These curves show diminishing returns, which is consistent with later applications of $\mathcal{R}_{\theta}$ appearing in fewer training examples. With the one-loop override, the marginal loop-count probabilities during mixed-depth training are $p_1=0.325$ and $p_2=p_3=p_4=0.225$. An example reaches loop $j$ only when $k\geq j$, so loops two, three, and four occur in 67.5\%, 45\%, and 22.5\% of mixed-depth examples, giving a mean loop count of $\bar{k}=2.35$ (Eq.~\ref{eq:elr_one_loop_marginal}).

\paragraph{Inference cost.} Because a one-loop forward pass executes each unique layer exactly once, ELR at $k=1$ matches the TTFT and per-token decode latency of its non-recurrent baseline. Each additional loop repeats only the recurrent block, reaching $2.33\times$ baseline TTFT and $2.06\times$ baseline decode latency at four loops for GDP-Hybrid-4B. Three loops therefore capture 96\% of the accuracy gain from one to four loops at 1.87$\times$ baseline TTFT. Quality--cost curves and measurement details are given in Appendix~\ref{appendix:inference_measurement}.

\begin{table}[t]
  \centering
  \caption{Cross-loop route replay on GDP-Hybrid-30B-A3B, relative to dynamically routed ELR (A). (B) applies replay to A's checkpoint at inference only while (C) is trained with replay. Validation loss and inference forward time are measured at $k=4$ on 8K-token sequences. Training update time is averaged over the mixed-depth training phase and profiled at fixed $k=4$.}
  \label{tab:route_replay}
  \footnotesize
  \setlength{\tabcolsep}{5pt}
  \begin{tabular}{@{}lcccc@{}}
    \toprule
    \multirow{2}{*}{Routing} & \multirow{2}{*}{\begin{tabular}[c]{@{}c@{}}$\Delta$ Val Loss \\ at $k=4$\end{tabular}}
      & \multicolumn{2}{c}{$\Delta$ Training Update Time}
      & \multirow{2}{*}{\begin{tabular}[c]{@{}c@{}}$\Delta$ Inference \\ Forward Time\end{tabular}} \\
    \cmidrule(lr){3-4}
     & & Mixed-depth & Fixed $k{=}4$ & \\
    \midrule
    (A) Dynamic routing       & \multicolumn{4}{c}{\textit{reference}} \\
    (B) Inference-only replay & $+0.0055$ & --        & --        & $-1.2\%$ \\
    (C) Replay training       & $+0.0038$ & $-11.7\%$ & $-19.6\%$ & $-1.2\%$ \\
    \bottomrule
  \end{tabular}
\end{table}

\subsection{Cross-Loop Route Replay}
\label{sec:experiments_route_replay}

Routing in GDP-Hybrid-30B-A3B ELR changes little across loops in aggregate but substantially per token (Table~\ref{tab:route_replay_diagnostics}). On 256 held-out 2K-token windows, per-expert selection counts at loops one and four have a cosine similarity of 0.978, averaged over the ten recurrent MoE layers, yet individual tokens retain only 53\% of their first-loop experts at loop four. To isolate the effect of this drift on each layer's computation, we give each recurrent MoE layer its loop-four input from the dynamically routed model and compute the routed-expert output twice: once with the experts and weights the router selects for that input, and once with those recorded for the same token at loop one. Because both computations share the same input, their difference reflects only the routing decision. The two outputs have a mean cosine similarity of 0.767. A control that shifts first-loop routes across positions within each document, so that each token receives another token's routes, gives 0.227. First-loop routes therefore retain much, but not all, of the token-specific routing information.

The resulting quality cost is small, while training time falls substantially (Table~\ref{tab:route_replay}). Inference-only replay (B) raises four-loop validation loss by 0.0055, and training with replay (C) narrows this to 0.0038 while leaving the ten-task average at $k=4$ unchanged within noise (70.90 versus 70.83 for A). Replay reduces router evaluations from 53 to 23 per four-loop forward pass but retains expert dispatch and computation, so inference forward time falls by only 1.2\% at 8K tokens. Training benefits more because replay also removes later-loop auxiliary-loss calculations and the backward work through repeated router evaluations. Update time falls by 11.7\% over the mixed-depth training phase and by 19.6\% at fixed $k=4$, with expert and all-to-all call counts unchanged. More details are given in Appendix~\ref{appendix:route_replay}.

\section{Koi-4B: A Post-Trained GDP-Hybrid ELR Model}
\label{sec:koi_4b}

\begin{figure}[t]
  \centering
  \includegraphics[width=0.84\linewidth]{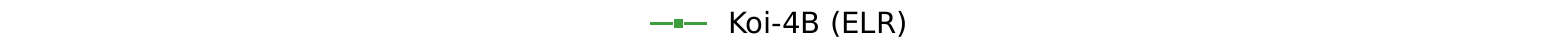}\par
  \vspace{-4pt}
  \begin{subfigure}[t]{0.46\linewidth}
    \centering
    \includegraphics[width=\linewidth]{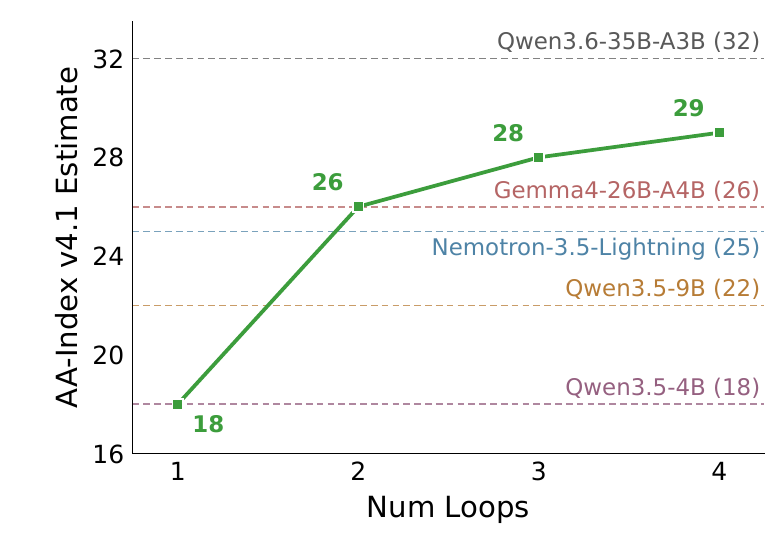}
    \caption{AA-Index v4.1 Estimate}
    \label{fig:koi_aa_index_scores}
  \end{subfigure}
  \hspace{0.05\linewidth}
  \begin{subfigure}[t]{0.44\linewidth}
    \centering
    \includegraphics[width=\linewidth]{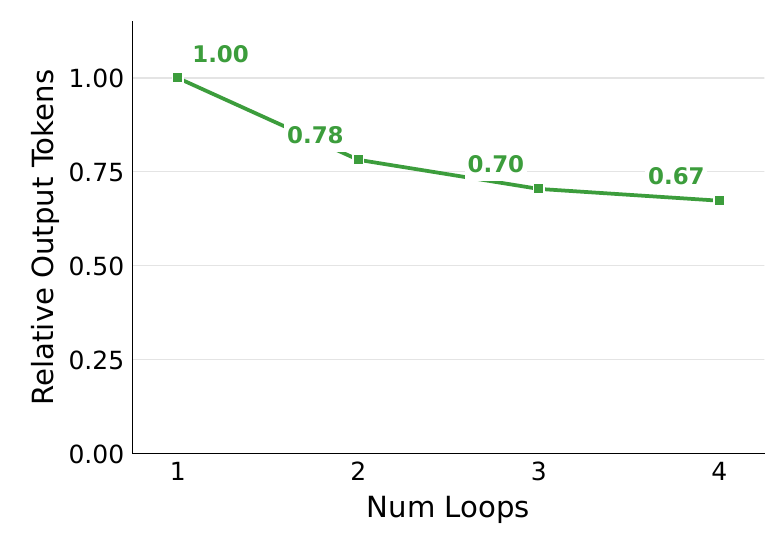}
    \caption{Relative Output Tokens}
    \label{fig:koi_output_tokens}
  \end{subfigure}
  \caption{Koi-4B evaluated at $k=1,\ldots,4$. (a) AA-Index v4.1 estimates, rounded to the nearest integer; dashed lines mark non-recurrent reference models. (b) Output tokens per rollout relative to $k=1$, as a geometric mean over eight of the AA-Index benchmarks (Appendix~\ref{appendix:koi_output_tokens}). All results come from our evaluation harness, not official Artificial Analysis results.}
  \label{fig:koi_aa_index}
\end{figure}

\subsection{Experimental Setup}
\label{sec:koi_experiment_setup}

\paragraph{Model and Training.} Koi-4B applies the full recipe of Section~\ref{sec:method} to the GDP-Hybrid-4B architecture of Section~\ref{sec:experiments}, adding loop-conditioned modulation (Section~\ref{sec:method_loop_conditioning}). We chose the dense hybrid because it targets the setting where elastic depth matters most. On a device with a fixed memory budget, a 4B model fits where a 30B-total mixture of experts does not, and looping lets the model that fits spend compute to close the gap. Post-training the MoE variant is left for future work. We pretrain it for 7T tokens at an 8K sequence length on a subset of the Nemotron 3 Ultra pretraining corpus~\citep{nvidia2026nemotron3ultra}, using one loop for the first half and mixed-depth training for the second half (Appendix~\ref{appendix:koi_training_details}). We then extend the context length with a 50B-token mid-training stage at a 256K sequence length, and post-train on the text-only data from Nemotron 3.5 Super~\citep{nvidia2026nemotronposttrainingv3} with four epochs of supervised fine-tuning (SFT) followed by 200 steps of reinforcement learning from verifiable rewards (RLVR). All stages after the switch use mixed-depth training. Pretraining, mid-training, and SFT sample loop counts uniformly with a 10\% one-loop rank override, while RLVR samples with a 4:1:1:1 ratio, favoring one-loop rollouts, which are faster to generate. Unlike Section~\ref{sec:experiments}, this run has no matched non-recurrent baseline, so its one-loop accuracy is compared to open models of similar size rather than to a model trained on the same data.

\paragraph{Evaluation.} We evaluate the single post-trained Koi-4B checkpoint at $k\in\{1,2,3,4\}$ on the benchmark suite and weighting of the Artificial Analysis Intelligence Index v4.1~\citep{artificialanalysis2026index41}: GDPval-AA v2 (Elo; 20\%), Terminal-Bench 2.1 (16\%), $\tau^3$-Bench Banking (14\%), Humanity's Last Exam (12\%), AA-Omniscience accuracy (8\%), SciCode (8\%), GPQA Diamond (6\%), AA-LCR (6\%), CritPt (6\%)\footnote{We do not run CritPt since models of this size score near zero.}, and AA-Omniscience non-hallucination (4\%). We also report IFBench~\citep{pyatkin2025ifbench}, LiveCodeBench v6 (pass@1)~\citep{jain2025livecodebench}, and MATH-500~\citep{lightman2024verify}, which are outside the v4.1 index. All scores, for Koi-4B and the reference models alike, come from our own harness rather than from Artificial Analysis and may differ from published figures. Koi-4B has no matched non-recurrent baseline, so we compare it with open models of similar size and larger.

\begin{table}[t]
  \caption{Koi-4B with per-loop and last-loop KV caches. $\Delta$ Score is last-loop minus per-loop at $k=4$, in percentage points, computed before rounding. Output Tokens is the ratio of mean output tokens per rollout, last-loop to per-loop, at $k=4$. Results at $k=2,3$ are in Appendix~\ref{appendix:koi_kv_results}.}
  \label{tab:koi_kv_cache}
  \centering
  \small
  \setlength{\tabcolsep}{4pt}
  \begin{tabular}{@{}lccccc@{}}
    \toprule
    \multirow{2}{*}{Benchmark}
      & \multicolumn{2}{c}{Per-loop}
      & Last-loop
      & \multirow{2}{*}{\shortstack{$\Delta$\\[1pt]Score}}
      & \multirow{2}{*}{\shortstack{Output\\[1pt]Tokens}} \\
    \cmidrule(lr){2-3}\cmidrule(lr){4-4}
      & $k=1$ & $k=4$ & $k=4$ & & \\
    \midrule
    Terminal-Bench 2.1 & 12.1 & 28.0 & 27.4 & $-0.6$ & $0.80\times$ \\
    $\tau^3$-Bench Banking & 11.3 & 21.8 & 21.8 & $\phantom{-}0.0$ & $0.86\times$ \\
    Humanity's Last Exam & 5.9 & 11.9 & 10.5 & $-1.4$ & $0.93\times$ \\
    AA-Omniscience accuracy & 8.1 & 9.3 & 8.5 & $-0.7$ & $0.82\times$ \\
    SciCode & 17.7 & 33.4 & 32.0 & $-1.4$ & $0.97\times$ \\
    GPQA Diamond & 59.9 & 72.8 & 72.2 & $-0.6$ & $0.99\times$ \\
    AA-LCR & 51.3 & 68.3 & 65.3 & $-3.0$ & $0.93\times$ \\
    AA-Omniscience non-halluc. & 86.1 & 85.9 & 88.2 & $+2.3$ & $0.82\times$ \\
    IFBench & 61.2 & 71.8 & 71.5 & $-0.3$ & $0.98\times$ \\
    LiveCodeBench v6 & 50.5 & 71.9 & 70.2 & $-1.6$ & $0.98\times$ \\
    MATH-500 & 88.8 & 97.2 & 97.0 & $-0.2$ & $1.01\times$ \\
    \midrule
    Attention KV memory (rel.\ to $k=1$) & $1\times$ & $2.5\times$ & $1\times$ & & \\
    \bottomrule
  \end{tabular}
\end{table}

\subsection{Inference-Time Depth Scaling}
\label{sec:koi_benchmark_results}

Koi-4B's AA-Index v4.1 estimate rises with every additional loop, from 18 at $k=1$ to 26, 28, and 29 at $k=2,3,4$ (Figure~\ref{fig:koi_aa_index}). At one loop, it's on par with Qwen3.5-4B, an open model of the same size. The second loop contributes 8 of the 11 points gained from one to four loops, mirroring the diminishing returns in Section~\ref{sec:experiments_loop_parity_gains}. Two loops already place Koi-4B above Qwen3.5-9B and Nemotron-3.5-Lightning and level with Gemma4-26B-A4B. At four loops, it scores 3 points above Gemma4-26B-A4B, which has about $6.5\times$ as many total parameters, and trails only Qwen3.6-35B-A3B, which has nearly $9\times$ as many total parameters. Each additional loop adds inference compute but no parameters, and Section~\ref{sec:koi_kv_cache} shows that attention KV memory can also stay at its one-loop level. Per-benchmark scores are given in Appendix~\ref{appendix:koi_results}.

For the same prompts, the same checkpoint produces shorter responses at higher loop counts while scoring higher (Figure~\ref{fig:koi_output_tokens}). Relative to one loop, output tokens per rollout fall by 21.8\%, 29.5\%, and 32.7\% at $k=2,3,4$ (geometric mean over benchmarks), and as with accuracy, the second loop accounts for most of the change. Additional loops thus reach higher scores with shorter reasoning traces, consistent with the model substituting latent depth for explicit chain-of-thought tokens. The shorter generations partly offset the added per-token cost. At two loops, per-token decode latency is about 1.36× that of one loop (Appendix~\ref{appendix:inference_measurement}), but rollouts are 0.78× as long, so decode time per rollout rises by only about 6\%. At four loops the figures are 2.06× and 0.67×, or about 1.4× per rollout (Appendix~\ref{appendix:koi_output_tokens}).

\subsection{Last Loop KV-Cache Inference}
\label{sec:koi_kv_cache}

Standard inference keeps a separate attention KV cache for each loop, so memory grows with depth. In Koi-4B, three of the six attention layers sit in the recurrent block, so at $k=4$ per-loop caching needs $2.5\times$ the one-loop KV memory, or 7.5 GiB against 3.0 GiB per 128K-token request in BF16. Last-loop KV inference instead shares one cache across loops. Each loop overwrites the cache, and subsequent tokens read earlier tokens' final-loop keys and values. The scheme requires no retraining and keeps attention KV memory at its one-loop size at every $k$. GDP layers keep per-loop state, which does not grow with sequence length and stays below 25 MiB per request at $k=4$.

Last-loop KV preserves most of the gain from additional loops (Table~\ref{tab:koi_kv_cache}). At four loops, score reductions relative to per-loop caching range from 0--3\%, while AA-Omniscience non-hallucination improves. Last-loop KV also produces fewer output tokens on nine of ten benchmarks, with much of the reduction coming from fewer requests reaching the output cap.

Koi-4B can therefore use additional inference depth with little increase in cache memory, at a small, benchmark-dependent accuracy cost. Additional details and results are in Appendix~\ref{appendix:koi_kv_results}.

\FloatBarrier
\section{Ablations}
\label{sec:ablations}

\begin{figure}[!t]
  \centering
  \begin{subfigure}[t]{0.52\linewidth}
    \centering
    \includegraphics[width=\linewidth]{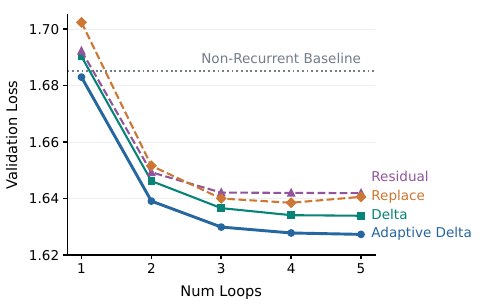}
    \caption{State Update Rules}
    \label{fig:ablation_state_update_rule}
  \end{subfigure}
  \hfill
  \begin{subfigure}[t]{0.42\linewidth}
    \centering
    \includegraphics[width=0.92\linewidth]{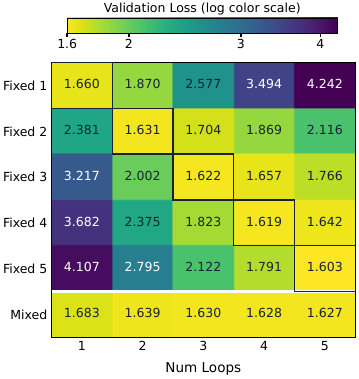}
    \caption{Fixed versus Mixed Depth Training}
    \label{fig:ablation_fixed_mixed}
  \end{subfigure}
  \caption{Architectural and training ablations on GPT-0.5B. (a) State update rules: solid lines mark rules with learned step sizes, and dashed lines mark parameter-free rules. (b) Fixed-depth versus mixed-depth training: outlined cells mark matching training and evaluation depths.}
  \label{fig:ablation_state_update_and_training}
\end{figure}

We isolate the architectural and training choices of ELR on GPT-0.5B, a dense Transformer trained for 300B tokens on a subset of the pretraining data used for Nemotron 3 Ultra~\citep{nvidia2026nemotron3ultra}, and report validation loss. Unless stated otherwise, each ELR variant trains with mixed-depth training from scratch with $K=5$ and a 15\% one-loop rank override, giving $p_1 = 0.32$ (Eq.~\ref{eq:elr_one_loop_marginal}).\footnote{GPT-0.5B omits input re-injection ($\tilde{s}_i = s_{i-1}$). Re-injection changes losses by at most 0.005 and leaves the ordering of every ablation in this section unchanged (see Appendix~\ref{appendix:ablation_input_injection}).}

\subsection{Adaptive Delta versus Alternative State Updates}
\label{sec:ablations_state_update_rule}

Adaptive delta is the only state update rule that matches the baseline at $k=1$, and it achieves the lowest loss at every loop count (Figure~\ref{fig:ablation_state_update_rule}). We compare it (Eq.~\ref{eq:elr_adaptive_delta_rule}) with two parameter-free rules, replace ($s_i = r_i$) and residual ($s_i = s_{i-1} + r_i$), and with delta, which substitutes a learned, input-independent per-channel rate $\alpha$, initialized to $0.05$, for the input-dependent gate $\beta_i$. Delta ranks second at every depth but trails adaptive delta by 0.006--0.007, indicating that conditioning the step size on the input is better than a fixed per-channel rate. The parameter-free rules perform worst. Replace has the highest one-loop loss and its loss rises from four to five loops, while residual stops improving after three loops.

\subsection{Fixed-Depth versus Mixed-Depth Training}
\label{sec:ablations_fixed_mixed_training}

Mixed-depth training makes a single model usable across multiple loop counts during inference (Figure~\ref{fig:ablation_fixed_mixed}). A model trained at one fixed loop count reaches the lowest loss at that count and degrades at every other $k$, with the gap growing with distance from its training depth in either direction. Because each fixed-depth model outperforms the mixed-depth model at its own depth, it serves as an empirical upper bound on the quality attainable at that depth. The mixed-depth model instead improves monotonically from one to five loops and trades this per-depth optimum for elasticity: a single checkpoint serves all five loop counts in place of five separately trained models, at some cost in loss at each depth.

The model trained only at $k=1$ reaches a lower loss than the non-recurrent baseline (1.660 versus 1.685), although both execute each unique layer once. Without input re-injection, the two differ only in the gate, which at one loop is a learned skip around the recurrent block, so the gate appears to help even without recurrence.

\begin{table}[t]
  \caption{Loop sampling schemes on GPT-0.5B (validation loss). Weighted sampling spreads the probability remaining after $p_1$ uniformly over $k\in\{2,\ldots,5\}$, so rows with the same $p_1$ share the same marginal distribution over $k$. Depths/Update is the maximum number of distinct loop counts contributing gradients to one update. Step time is relative to synchronized uniform sampling.}
  \label{tab:ablation_sampling}
  \centering
  \small
  \setlength{\tabcolsep}{4pt}
  \begin{tabular}{llccccccc}
    \toprule
    $p_1$ & Sampling & Depths/Update & Step Time & $k=1$ & $k=2$ & $k=3$ & $k=4$ & $k=5$ \\
    \midrule
    -- & Non-recurrent baseline & -- & -- & 1.6852 & -- & -- & -- & -- \\
        \midrule
    \multirow{2}{*}{0.20} & Synchronized, uniform & 1        & $1.00\times$ & 1.7077 & 1.6473 & 1.6344 & 1.6318 & 1.6313 \\
                          & Independent, uniform  & $\leq5$  & $1.36\times$ & 1.6937 & \textbf{1.6330} & \textbf{1.6193} & \textbf{1.6157} & \textbf{1.6147} \\
    \midrule
    \multirow{3}{*}{0.32} & Synchronized, weighted       & 1       & $1.00\times$ & 1.6871 & 1.6436 & 1.6343 & 1.6320 & 1.6316 \\
                          & Synchronized, 15\% override  & $\leq2$ & $1.00\times$ & 1.6830 & 1.6391 & 1.6299 & 1.6278 & 1.6273 \\
                          & Independent, weighted        & $\leq5$ & $1.36\times$ & \textbf{1.6774} & 1.6334 & 1.6231 & 1.6200 & 1.6194 \\
    \bottomrule
  \end{tabular}
\end{table}

\Needspace{8\baselineskip}
\subsection{Synchronized versus Independent-Rank Loop Sampling}
\label{sec:ablations_sampling}

The sampling scheme determines how many depths contribute gradients to each optimizer update. Under synchronized sampling, all data-parallel ranks run the same loop count in a step, so each update covers one depth, or at most two with one-loop overrides. Under independent-rank sampling, each rank draws its own loop count, so one update can cover every depth, but the step waits for the rank with the largest count. At $p_1=0.32$, the three schemes in Table~\ref{tab:ablation_sampling} share the same marginal distribution over $k$, with $p_k=0.17$ for $k\geq2$, and differ only in how depths are grouped within updates. Loss decreases as more depths share an update. One-loop loss falls from 1.6871 under synchronized weighted sampling to 1.6830 with 15\% overrides and 1.6774 under independent-rank sampling, and the same ordering holds at every $k\geq2$. Independent-rank sampling gives the lowest loss but increases step time by about 36\%. One-loop overrides recover part of this gain at synchronized step time so we use them in our main experiments.

\subsection{One-Loop Rank Overrides}
\label{sec:ablations_one_loop_rank_override}

Under synchronized sampling, one-loop rank overrides are needed for one-loop parity (Figure~\ref{fig:ablation_override}). Without overrides ($f=0$), uniform sampling reaches one-loop loss +0.0225 above the baseline. One-loop loss decreases monotonically as $f$ increases, and $f=0.15$ ($p_1=0.32$) is the smallest tested fraction that falls below the baseline. Larger fractions buy further one-loop gains at the cost of deeper loops, which motivates choosing $f$ so that $p_1$ is approximately one third (Section~\ref{sec:method_mix_training_one_loop_override}). The benefit does not come from the higher one-loop marginal alone, since synchronized weighted sampling at the same $p_1$ without overrides remains above the baseline (Section~\ref{sec:ablations_sampling}).

\subsection{Delayed Mixed-Depth Training}
\label{sec:ablations_delayed_mix_depth_training}

\begin{figure}[!t]
  \centering
  \includegraphics[width=0.84\linewidth]{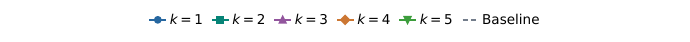}\par
  \vspace{-8pt}
  \begin{subfigure}[t]{0.4\linewidth}
    \centering
    \includegraphics[width=\linewidth]{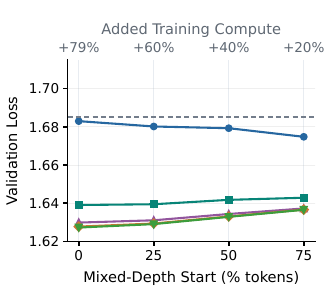}
    \caption{Delayed Mixed-Depth Training}
    \label{fig:ablation_delayed}
  \end{subfigure}
  \hspace{0.1\linewidth}
  \begin{subfigure}[t]{0.4\linewidth}
    \centering
    \includegraphics[width=\linewidth]{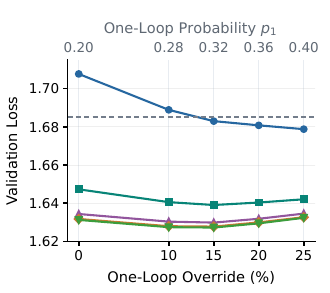}
    \caption{One-Loop Override}
    \label{fig:ablation_override}
  \end{subfigure}
  \caption{Training choices on GPT-0.5B. Each x-axis point is a separately trained model, and each line shows validation loss at one inference loop count $k$ across these models. The dashed line marks the non-recurrent baseline (1.685). (a) Point at which training switches from one loop to mixed depth ($f=0.15$); 0\% is mixed-depth training from scratch, and added compute is estimated from Eq.~\ref{eq:delayed_training_cost}. (b) One-loop override fraction $f$, with mixed-depth training from scratch.}
  \label{fig:ablation_training}
\end{figure}

Delaying mixed-depth training retains most of the gain from additional loops at a fraction of its training cost (Figure~\ref{fig:ablation_delayed}). Mixed-depth training from scratch costs an estimated $1.79\times$ the baseline's training compute (Eq.~\ref{eq:delayed_training_cost}). Starting it after 25\%, 50\%, or 75\% of training tokens lowers this to $1.60\times$, $1.40\times$, and $1.20\times$, while retaining 97\%, 90\%, and 84\% of the five-loop improvement over the baseline. At the 75\% start, the estimated added training compute falls from about 79\% to 20\% for a 0.009 increase in five-loop loss, and one-loop loss stays below the baseline at every start point. We therefore use the 75\% start for our main experiments (Section~\ref{sec:experiments_setup}).

\subsection{Loop-Conditioned Modulation}
\label{sec:ablations_loop_conditioning}

\begin{figure}[!ht]
  \centering
  \includegraphics[width=0.43\linewidth]{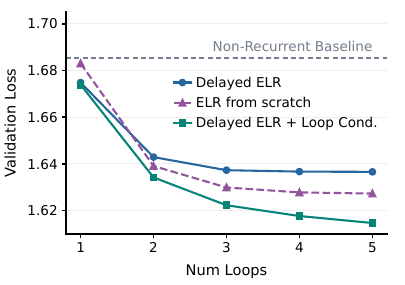}
  \caption{Loop-conditioning on GPT-0.5B. Delayed models switch to mixed depth after 75\% of tokens, and the from-scratch model trains at mixed depth throughout.}
  \label{fig:ablation_conditioning}
\end{figure}

Loop-conditioned modulation leaves one-loop loss nearly unchanged and lowers loss at every additional loop, with gains growing with loop count (Figure~\ref{fig:ablation_conditioning}). We add the modulation of Section~\ref{sec:method_loop_conditioning} to the delayed model that switches to mixed depth after 75\% of training tokens, introducing it at the switch with all heads initialized to zero. Loop-conditioned modulation adds 31.6M parameters (6.34\% of the GPT-0.5B backbone). One-loop loss moves from 1.6748 to 1.6739, while loss falls by 0.009 at $k=2$ and by 0.022 at $k=5$, increasing the loss reduction from one to five loops from 0.038 to 0.059. The conditioned model also achieves lower validation-loss point estimates at every loop count than unconditioned mixed-depth training from scratch (1.6147 versus 1.6273 at five loops), despite using mixed depths only during the final quarter of training.

\section{Related Work}
\label{sec:related_work}

\paragraph{Elastic recurrent depth.}
Huginn~\citep{geiping2025latent} and Ouro~\citep{zhu2025scaling} train looped language models whose depth can vary at inference, Mixture-of-Recursions routes tokens to learned recursion depths~\citep{bae2025mixture}, and LoopFormer conditions a shared stack on normalized time and step size so that shortcut trajectories align with the full-depth trajectory~\citep{jeddi2026loopformer}. LoopMoE conditions each iteration on its index and the token's hidden state through an adaptive layer norm~\citep{chen2026loopmoe}, and Huginn reports that conditioning the recurrent block on the step index hurt its ability to extrapolate beyond the trained depth~\citep{geiping2025latent}. Our conditioning uses the same signal as LoopFormer, progress and budget, but is grafted onto a trained one-loop checkpoint with zero-initialized heads, so the conditioned model starts as that checkpoint and the deeper passes learn to depart from it; we do not evaluate extrapolation. LoopFormer's dense comparisons match either unique parameters or block-level inference FLOPs but not both. During training, Huginn samples a single recurrence depth per step, shared by all data-parallel workers~\citep{geiping2025latent}, whereas the one-loop rank override mixes one-loop and deeper gradients within a single update. Closest to our goal, stochastic RINS drops recursion rounds during training and, with depth-specific linear adapters, matches or exceeds a training-compute-matched baseline at one recursion in 600M-parameter models~\citep{alabdulmohsin2025recursive}. That result is reported in perplexity, and without the adapters, stronger one-recursion protection reduces gains at higher depths. ELR reaches one-loop parity without depth-specific parameters, with downstream evaluation at 4B dense-hybrid and 30B-A3B MoE scale.

\paragraph{Regulating recurrent dynamics.}
Hyperloop Transformers apply matrix-valued hyper-connections after each loop~\citep{zeitoun2026hyperloop}, Parcae constrains input-injection dynamics with stability conditions derived from a linearized system~\citep{prairie2026parcae}, STARS regularizes the Jacobian spectral radius~\citep{yang2026stars}, and Training-Free Looped Transformers reinterpret repeated frozen-model computation as damped substeps~\citep{chen2026trainingfree}. ELR's adaptive-delta update starts from a small step size but learns it end to end without explicit stability constraints, and is designed to match the non-recurrent baseline at one loop while improving with additional loops.

\paragraph{Acquiring recurrence late.}
Retrofitted Recurrence~\citep{mcleish2025retrofitted} and Relaxed Recursive Transformers~\citep{bae2025relaxed} convert pretrained non-recurrent models into recurrent ones, using a recurrence curriculum and depth-specific low-rank adapters, respectively. Our delayed mixed-depth training instead continues a model that is recurrent by construction, changing only the loop-count distribution, with no architectural change or optimizer restart.

\paragraph{Recurrent MoE and serving.}
Sparse Universal Transformer~\citep{tan2023sparse} and LoopMoE~\citep{chen2026loopmoe} combine sparse expert routing with weight-shared depth. Loopie trains looped MoE models at up to 20B total parameters with a fixed two recurrent steps and asks whether looping beats parameter scaling at matched compute~\citep{gao2026loop}; we instead take the non-recurrent MoE as given and ask what it costs to make it elastic. Read-ME decouples routing from a non-recurrent MoE backbone so that expert decisions can be precomputed for scheduling~\citep{cai2024readme}, whereas cross-loop route replay reuses a looped model's own first-loop routes on later loops. Continuous Depth Batching schedules tokens that exit at different loop counts~\citep{schwethelm2026continuous}, whereas we serve each request at a fixed loop count.

\section{Conclusion}
\label{sec:conclusion}

Elastic Latent Recurrence shows that a looped model can serve a single checkpoint across a range of inference depths without paying a one-loop tax. Three choices make this work. A gated adaptive-delta update and within-step mixing of loop counts preserve the one-loop endpoint, and conditioning each pass on its loop index and budget lets passes specialize, lowering loss at every added loop. At one loop, an ELR model executes every unique layer exactly once; trained on the same data as its non-recurrent baseline, it is 0.5 points below the baseline at 4B and 1.0 point above it at 30B-A3B, and four loops of the same checkpoint add 7.25 and 4.75 points. Elastic depth need not be trained from scratch. Introducing mixed depths in the final quarter of pretraining costs about 1.16× the baseline's training compute, and conditioning grafted at the switch with zero-initialized heads makes this late schedule outperform mixed-depth training from the start. Cross-loop route replay cuts mixed-depth update time in recurrent MoE models by 11.7\%, and last-loop KV caching holds attention memory at the one-loop level at every depth. Carried through pretraining and post-training, the recipe yields Koi-4B, a single checkpoint that scores 18 on the Artificial Analysis Intelligence Index at one loop and 29 at four loops in our harness.

Several questions remain open. Gains diminish after the second loop, which is consistent with later loops appearing in fewer training examples, but we have not tested whether reweighting the loop-count distribution recovers them. Conditioning closes most of the gap to fixed-depth specialists at intermediate depths but not at the endpoints, and we have not isolated why. Our serving setting fixes the loop count per request; combining ELR with per-request or per-token adaptive depth, which budget conditioning complicates, and post-training the MoE variant are the next steps.

\bibliography{references}
\bibliographystyle{references}

\appendix
\newpage
\section{Model Architecture and Training Details}
\label{appendix:model_training_details}

All models are implemented in Megatron-LM. Each self-attention, feed-forward, GDP, and MoE layer counts as one layer. The ELR models in Section~\ref{sec:experiments} and the GPT-0.5B ablations share their backbone configurations with their non-recurrent baselines. Without loop-conditioned modulation, ELR adds only the adaptive-delta gate, whose $d^2+2d$ parameters amount to about 0.2\% of the baseline parameters for GPT-0.5B and GDP-Hybrid-4B, and 0.02\% of the total (0.2\% of the active) parameters for GDP-Hybrid-30B-A3B. The loop-conditioned GPT-0.5B ablation uses conditioning dimension 256, low-rank dimension $r=256$, and maximum loop count $K=5$. Koi-4B additionally uses loop-conditioned modulation and has no matched non-recurrent baseline; its architecture and training stages are described in Appendix~\ref{appendix:koi_training_details}.

\subsection{Architectures}
\label{appendix:architectures}
GPT-0.5B is a dense Transformer whose 30 layers alternate between self-attention and SwiGLU MLP layers, and its ELR variant partitions them 8/14/8. GDP-Hybrid-4B and GDP-Hybrid-30B-A3B follow the Nemotron 3 Nano hybrid layout~\citep{nvidia2025nemotron3nano}, with Mamba-2 layers replaced by Gated DeltaProduct (GDP) layers~\citep{siems2025deltaproduct}. Both use the same layer ordering, written below with G for GDP, A for attention, and F for a feed-forward layer, which is a dense MLP in GDP-Hybrid-4B and an MoE layer in GDP-Hybrid-30B-A3B. Bars mark the 14/24/14 ELR partition:
\begin{center}
\footnotesize
\texttt{GFGFGAFGFGFGAF\,|\,GFGFGAFGFGFGAFGFGFGAFGFG\,|\,FGFGAFGFGFGFGF}
\end{center}
All models use RMSNorm, linear layers without bias, untied embeddings, no dropout, BF16 training, and a 131,072-token vocabulary. Table~\ref{tab:appendix_architecture} lists the remaining dimensions.

\begin{table}[h]
  \caption{Backbone configurations, shared by each ELR model and its non-recurrent baseline. Koi-4B uses the GDP-Hybrid-4B configuration. Partial RoPE rotates 50\% of each attention head's dimensions.}
  \label{tab:appendix_architecture}
  \centering
  \small
  \setlength{\tabcolsep}{5pt}
  \begin{tabularx}{\linewidth}{@{}>{\raggedright\arraybackslash}Xccc@{}}
    \toprule
    & GPT-0.5B & GDP-Hybrid-4B & GDP-Hybrid-30B-A3B \\
    \midrule
    Layers (GDP / attention / FF) & 0 / 15 / 15 & 23 / 6 / 23 & 23 / 6 / 23 \\
    ELR partition ($\mathcal{I}$ / $\mathcal{R}_{\theta}$ / $\mathcal{O}$) & 8 / 14 / 8 & 14 / 24 / 14 & 14 / 24 / 14 \\
    Feed-forward type & Dense & Dense & MoE \\
    Activation & SwiGLU & Squared ReLU & Squared ReLU \\
    Hidden size $d$ & 1,024 & 3,072 & 2,688 \\
    FF intermediate size & 3,968 & 12,288 & 1,856 per expert \\
    Query heads / KV groups & 8 / 4 & 24 / 8 & 32 / 2 \\
    Attention head dim. & 128 & 128 & 128 \\
    Position encoding & RoPE ($10^6$) & None & Partial RoPE ($10^5$) \\
    GDP heads / head dim.\ / state dim. & -- & 96 / 32 / 64 & 16 / 64 / 128 \\
    Routed experts / top-$k$ & -- & -- & 512 / 22 \\
    Expert latent dim.\ / shared expert size & -- & -- & 672 / 3,712 \\
    \bottomrule
  \end{tabularx}
\end{table}

\paragraph{MoE routing.}
GDP-Hybrid-30B-A3B computes sigmoid router scores in FP32. It balances expert load with a learned per-expert score bias and a sequence-level auxiliary loss with coefficient $10^{-4}$.

\subsection{Optimization}
\label{appendix:optimization}
Table~\ref{tab:appendix_training} lists the optimization settings for the matched experiments and GPT-0.5B ablations, which each baseline shares with its ELR counterpart. These models train on a subset of the Nemotron 3 Ultra pretraining data~\citep{nvidia2026nemotron3ultra}. GDP-Hybrid-30B-A3B applies Muon to two-dimensional weight matrices, with momentum 0.9, 16 Newton--Schulz iterations using Polar Express coefficients, and spectral update scaling with an extra factor of 0.2. AdamW updates all other parameters. For these runs, the learning rate follows a warmup--stable--decay schedule with a $1-\sqrt{\cdot}$ decay to the minimum learning rate over the final 15\% of training tokens.

\begin{table}[h]
  \caption{Optimization and ELR training settings. Each non-recurrent baseline uses the same data, token budget, and optimization settings as its ELR counterpart. $p_1$ and $\bar{k}$ are nominal values from Eq.~\ref{eq:elr_one_loop_marginal}.}
  \label{tab:appendix_training}
  \centering
  \small
  \setlength{\tabcolsep}{5pt}
  \begin{tabular}{@{}lccc@{}}
    \toprule
    & GPT-0.5B & GDP-Hybrid-4B & GDP-Hybrid-30B-A3B \\
    \midrule
    Training tokens & 300B & 1T & 300B \\
    Optimizer & AdamW & AdamW & Muon + AdamW \\
    AdamW $(\beta_1, \beta_2)$ / weight decay & (0.9, 0.95) / 0.1 & (0.9, 0.95) / 0.1 & (0.9, 0.95) / 0.1 \\
    Peak / minimum learning rate & $1.8{\times}10^{-3}$ / $1.8{\times}10^{-5}$ & $1.2{\times}10^{-3}$ / $1.2{\times}10^{-5}$ & $1.2{\times}10^{-3}$ / $1.2{\times}10^{-5}$ \\
    Warmup tokens & 8.4B & 33.6B & 8.4B \\
    Sequence length / global batch & 8,192 / 768 & 8,192 / 768 & 8,192 / 768 \\
    Gradient clipping & 1.0 & 1.0 & 1.0 \\
    Initialization std & 0.028 & 0.0162 & 0.0173 \\
    \midrule
    Maximum loop count $K$ & 5 & 4 & 4 \\
    One-loop override fraction $f$ & 0.15 & 0.10 & 0.10 \\
    $p_1$ / $\bar{k}$ & 0.32 / 2.70 & 0.325 / 2.35 & 0.325 / 2.35 \\
    Mixed-depth phase & All tokens & Final 25\% & Final 25\% \\
    \bottomrule
  \end{tabular}
\end{table}

\subsection{Mixed-Depth Training}
\label{appendix:mixed_depth_training}
A single majority loop count is drawn for each microbatch and shared by all data-parallel ranks. Of the $N_{\mathrm{DP}}$ ranks, $\operatorname{round}(fN_{\mathrm{DP}})$ contiguous ranks run one loop, and the starting index of this group shifts on every microbatch. Gradients are backpropagated through all executed loops without truncation.

In delayed mixed-depth training, the continuation restores the model weights, optimizer state, and learning-rate schedule from the one-loop checkpoint. For GDP-Hybrid-30B-A3B, the dynamically routed and route-replay ELR models start from the same one-loop checkpoint. The fixed-depth GPT-0.5B models in Section~\ref{sec:ablations_fixed_mixed_training} train without overrides.

\FloatBarrier
\subsection{Koi-4B Architecture and Training}
\label{appendix:koi_training_details}

\paragraph{Architecture.}
Koi-4B uses the GDP-Hybrid-4B backbone and 14/24/14 partition in Table~\ref{tab:appendix_architecture}, with the ELR architecture of Section~\ref{sec:method_elr}. It adds the loop-conditioned modulation of Section~\ref{sec:method_loop_conditioning} with conditioning dimension 256, low-rank dimension $r=256$, and $K=4$, adding 155.6M parameters (3.94\% of the backbone).

\paragraph{Pretraining and context extension.}
Koi-4B is pretrained for 7T tokens at sequence length 8,192 on a subset of the Nemotron 3 Ultra corpus~\citep{nvidia2026nemotron3ultra}. The first 44\% of tokens (approximately 3.1T) use the diversity-focused Phase 1 blend, and the remaining 56\% (approximately 3.9T) use the quality-focused Phase 2 blend. Independently of this blend transition, the loop schedule switches at 50\% of the token budget: the first half uses $k=1$, and the second half draws the majority loop count uniformly from $k=1,\ldots,4$ with a rotating 10\% one-loop rank override, giving nominal $p_1=0.325$ and $\bar{k}=2.35$ (Eq.~\ref{eq:elr_one_loop_marginal}). The zero-initialized conditioning heads are introduced at this switch. By Eq.~\ref{eq:delayed_training_cost}, this schedule has an estimated layer-execution cost of $1.31\times$ that of pretraining at one loop throughout. A subsequent 50B-token mid-training stage extends the context length to 262,144 tokens with the same mixed-depth schedule. Table~\ref{tab:koi_training_stages} summarizes the training stages.

\paragraph{Pretraining optimization.}
Pretraining uses AdamW with $(\beta_1,\beta_2)=(0.9,0.95)$, weight decay 0.1, gradient clipping at 1.0, and a global batch size of 768 sequences. The learning rate warms up linearly over 33.6B tokens to $1.2\times10^{-3}$, remains constant, and follows a $1-\sqrt{\cdot}$ decay to $1.2\times10^{-5}$ over the final 15\% of training tokens.

\begin{table}[htbp]
  \caption{Koi-4B training stages. $p_1$ and $\bar{k}$ are nominal values from Eq.~\ref{eq:elr_one_loop_marginal} or the sampling ratio.}
  \label{tab:koi_training_stages}
  \centering
  \small
  \setlength{\tabcolsep}{4pt}
  \begin{tabularx}{\linewidth}{@{}l>{\raggedright\arraybackslash}X>{\raggedright\arraybackslash}X@{}}
    \toprule
    Stage & Data and budget & Loop schedule \\
    \midrule
    Pretraining & Nemotron 3 Ultra subset; 7T tokens at 8K context & One loop for the first 50\%; then uniform over $k=1,\ldots,4$ with a 10\% one-loop rank override ($p_1=0.325$, $\bar{k}=2.35$) \\
    Mid-training & Context extension; 50B tokens at 256K context & Same as the pretraining mixed-depth phase \\
    SFT & Nemotron 3.5 Super text-only data; four epochs & Same as the pretraining mixed-depth phase \\
    RLVR & Nemotron 3.5 Super text-only data; 200 steps & Sampling ratio $4:1:1:1$ for $k=1,2,3,4$ ($p_1\approx0.57$, $\bar{k}\approx1.86$) \\
    \bottomrule
  \end{tabularx}
\end{table}

\paragraph{Post-training.}
Koi-4B undergoes four epochs of supervised fine-tuning on the text-only Nemotron 3.5 Super data~\citep{nvidia2026nemotronposttrainingv3}, followed by 200 steps of reinforcement learning from verifiable rewards. SFT uses packed sequences of length 256K, a global batch size of 48, and AdamW with $(\beta_1,\beta_2)=(0.9,0.95)$, weight decay 0.1, and gradient clipping at 1.0. The learning rate is $2\times10^{-5}$ after a 50-step warmup and remains constant thereafter. SFT uses the same mixed-depth settings as pretraining. RLVR uses asynchronous GRPO with 16 rollouts per prompt, a batch of 256 prompts (4,096 rollouts per step), and a constant learning rate of $4\times10^{-6}$ (Adam, no weight decay, gradient clipping at 1.0). Advantages use a leave-one-out baseline with reward normalization. The loss is token-level, with no KL penalty and truncated importance sampling (weights clipped to $[0.2, 5]$) to correct for the one-step policy lag of asynchronous generation. Sequences whose generation and and training token probabilities differ by more than a factor of two on average are masked from the loss. Rollouts are sampled at temperature 1.0 with up to 24,576 new tokens in a 64K context. Loop counts are assigned per prompt: all 16 rollouts of a prompt use the same $k$, and prompts are split in a $4:1:1:1$ ratio across $k=1,2,3,4$, favoring the faster one-loop rollouts. A pass-rate curriculum routes harder prompts to higher $k$. Generation runs on separate inference engines, each serving a fixed loop count, with engines allocated $4:2:3:4$ across $k=1$–$4$ to balance the per-loop decode cost. We train for 200 steps.

\section{Matched-Baseline Benchmark Results}
\label{appendix:downstream_results}

\subsection{Evaluation Protocol}
\label{appendix:evaluation_protocol}
Evaluation uses task-specific decoding and scoring protocols. Every ELR checkpoint is evaluated at each loop count in its trained range, and the non-recurrent baseline uses the same protocol. Table~\ref{tab:benchmark_protocol} lists the benchmarks, few-shot counts, and metrics. GSM8K and MATH-500 answers are evaluated using mathematical equivalence via Math-Verify.

\begin{table}[!ht]
  \caption{Downstream evaluation benchmarks.}
  \label{tab:benchmark_protocol}
  \centering
  \footnotesize
  \setlength{\tabcolsep}{4pt}
  \begin{tabular}{@{}llcl@{}}
    \toprule
    Benchmark & Domain & Shots & Metric \\
    \midrule
    MMLU~\citep{hendrycks2021mmlu} & Knowledge & 5 & Accuracy \\
    MMLU-Pro~\citep{wang2024mmlupro} & Knowledge & 5 & Accuracy \\
    ARC-Challenge~\citep{clark2018arc} & Science QA & 25 & Accuracy \\
    WinoGrande~\citep{sakaguchi2021winogrande} & Commonsense & 5 & Accuracy \\
    CommonsenseQA~\citep{talmor2019commonsenseqa} & Commonsense & 7 & Accuracy \\
    RACE-high~\citep{lai2017race} & Reading comprehension & 0 & Accuracy \\
    GSM8K~\citep{cobbe2021gsm8k} & Math & 8 (CoT) & Answer accuracy \\
    MATH-500~\citep{lightman2024verify} & Math & 4 & pass@1 \\
    HumanEval~\citep{chen2021humaneval} & Code & 0 & pass@1 \\
    MBPP (sanitized)~\citep{austin2021mbpp} & Code & 3 & pass@1 \\
    \bottomrule
  \end{tabular}
\end{table}

\FloatBarrier
\subsection{GDP-Hybrid-4B}
\label{appendix:downstream_gdp_hybrid_4b}

Table~\ref{tab:downstream_gdp_hybrid_4b} reports the per-benchmark scores underlying Figure~\ref{fig:downstream_scaling_gdp_hybrid_4b}. All ELR columns come from a single checkpoint evaluated at $k=1,\ldots,4$.

\begin{table}[!ht]
  \centering
  \caption{Downstream benchmark scores (\%) for GDP-Hybrid-4B trained for 1T tokens.}
  \label{tab:downstream_gdp_hybrid_4b}
  \begingroup
  \footnotesize
  \setlength{\tabcolsep}{2pt}
  \renewcommand{\arraystretch}{1.18}
  \begin{tabular}{@{}cc>{\columncolor{black!6}}cccc@{}}
    \toprule
    \multirow{2}{*}{Benchmark} & \multirow{2}{*}{Baseline} & \multicolumn{4}{c}{ELR} \\
    \cmidrule(lr){3-6}
     &  & $k=1$ & $k=2$ & $k=3$ & $k=4$ \\
    \midrule
    MMLU & 62.1 & 62.5 & 66.7 & 67.2 & 67.6 \\
    MMLU-Pro & 41.3 & 39.5 & 45.0 & 47.0 & 47.3 \\
    ARC-Challenge & 79.5 & 79.8 & 83.8 & 84.6 & 85.0 \\
    WinoGrande & 70.8 & 73.5 & 74.8 & 75.2 & 75.6 \\
    CommonsenseQA & 71.4 & 68.8 & 75.5 & 77.2 & 77.8 \\
    RACE-high & 76.9 & 74.4 & 80.9 & 81.2 & 82.4 \\
    \addlinespace[3pt]
    GSM8K & 76.3 & 71.0 & 81.0 & 82.3 & 82.8 \\
    MATH-500 & 64.1 & 69.3 & 75.5 & 79.6 & 80.6 \\
    \addlinespace[3pt]
    HumanEval & 57.3 & 56.1 & 66.5 & 64.6 & 64.0 \\
    MBPP & 56.8 & 56.4 & 58.8 & 61.9 & 60.7 \\
    \midrule
    \textbf{Avg.} & \textbf{65.65} & \textbf{65.13} & \textbf{70.85} & \textbf{72.08} & \textbf{72.38} \\
    \bottomrule
  \end{tabular}
  \endgroup
\end{table}

\FloatBarrier
\subsection{GDP-Hybrid-30B-A3B}
\label{appendix:downstream_gdp_hybrid_30b_a3b}

Table~\ref{tab:downstream_gdp_hybrid_30b_a3b} reports the per-benchmark scores underlying Figure~\ref{fig:downstream_scaling_gdp_hybrid_30b_a3b} for ELR with dynamic routing and with cross-loop route replay (Section~\ref{sec:experiments_route_replay}). Both ELR models start the mixed-depth phase from the same one-loop checkpoint.

\begin{table}[!ht]
  \centering
  \caption{Downstream benchmark scores (\%) for GDP-Hybrid-30B-A3B trained for 300B tokens.}
  \label{tab:downstream_gdp_hybrid_30b_a3b}
  \begingroup
  \footnotesize
  \setlength{\tabcolsep}{2pt}
  \renewcommand{\arraystretch}{1.18}
  \begin{tabular}{@{}cc>{\columncolor{black!6}}cccc>{\columncolor{black!6}}cccc@{}}
    \toprule
    \multirow{2}{*}{Benchmark} & \multirow{2}{*}{Baseline} & \multicolumn{4}{c}{ELR} & \multicolumn{4}{c}{ELR + Route Replay} \\
    \cmidrule(lr){3-6}\cmidrule(lr){7-10}
     &  & $k=1$ & $k=2$ & $k=3$ & $k=4$ & $k=1$ & $k=2$ & $k=3$ & $k=4$ \\
    \midrule
    MMLU & 62.04 & 65.55 & 68.48 & 69.32 & 69.28 & 65.23 & 68.51 & 69.15 & 69.17 \\
    MMLU-Pro & 39.07 & 38.40 & 44.35 & 45.78 & 45.94 & 39.27 & 44.23 & 45.42 & 45.68 \\
    ARC-Challenge & 77.39 & 79.78 & 83.70 & 84.73 & 84.73 & 79.35 & 84.90 & 85.32 & 85.32 \\
    WinoGrande & 74.74 & 74.03 & 76.72 & 77.19 & 78.69 & 74.03 & 75.45 & 75.85 & 76.01 \\
    CommonsenseQA & 69.53 & 72.56 & 77.31 & 78.13 & 77.72 & 71.66 & 75.59 & 77.31 & 77.64 \\
    RACE-high & 75.12 & 77.89 & 83.54 & 83.64 & 83.06 & 79.14 & 82.87 & 84.78 & 84.50 \\
    \addlinespace[3pt]
    GSM8K & 79.68 & 80.44 & 86.13 & 86.35 & 86.96 & 79.91 & 85.06 & 86.05 & 86.81 \\
    MATH-500 & 60.78 & 63.34 & 65.84 & 65.90 & 65.95 & 66.53 & 70.47 & 68.51 & 68.54 \\
    \addlinespace[3pt]
    HumanEval & 49.39 & 45.73 & 50.00 & 47.56 & 50.61 & 43.29 & 49.39 & 51.83 & 50.00 \\
    MBPP & 62.65 & 63.04 & 66.15 & 65.37 & 65.37 & 61.09 & 63.04 & 65.37 & 65.37 \\
    \midrule
    \textbf{Avg.} & \textbf{65.04} & \textbf{66.08} & \textbf{70.22} & \textbf{70.40} & \textbf{70.83} & \textbf{65.95} & \textbf{69.95} & \textbf{70.96} & \textbf{70.90} \\
    \bottomrule
  \end{tabular}
  \endgroup
\end{table}

\FloatBarrier

\section{Koi-4B Evaluation and Detailed Results}
\label{appendix:koi_results}

Table~\ref{tab:koi_benchmark_results} reports per-benchmark scores for the single post-trained Koi-4B checkpoint at $k=1,\ldots,4$, under the evaluation setup of Section~\ref{sec:koi_experiment_setup}. All scores come from our evaluation harness. CritPt is not run, since models of this size score near zero on it, and contributes zero to the index at every loop count. Index estimates are given to two decimal places, and Figure~\ref{fig:koi_aa_index} rounds them to the nearest integer.

\begin{table}[!ht]
  \centering
  \caption{Koi-4B benchmark scores across inference loop counts. Parenthesized values are AA-Index v4.1 weights. GDPval-AA v2 is reported as Elo, and all other benchmarks as percentages.}
  \label{tab:koi_benchmark_results}
  \begingroup
  \small
  \setlength{\tabcolsep}{5pt}
  \renewcommand{\arraystretch}{1.15}
  \begin{tabular}{@{}lrrrr@{}}
    \toprule
    Benchmark (index weight) & $k=1$ & $k=2$ & $k=3$ & $k=4$ \\
    \midrule
    \textbf{AA-Index v4.1 estimate} & \textbf{17.52} & \textbf{25.57} & \textbf{27.93} & \textbf{29.23} \\
    \midrule
    GDPval-AA v2, Elo (20\%) & 611 & 899 & 960 & 997 \\
    Terminal-Bench 2.1 (16\%) & 12.1 & 22.9 & 26.5 & 28.0 \\
    $\tau^3$-Bench Banking (14\%) & 11.3 & 16.8 & 19.8 & 21.8 \\
    Humanity's Last Exam (12\%) & 5.9 & 9.1 & 10.1 & 11.9 \\
    AA-Omniscience accuracy (8\%) & 8.1 & 8.8 & 8.6 & 9.3 \\
    SciCode (8\%) & 17.7 & 28.0 & 30.8 & 33.4 \\
    GPQA Diamond (6\%) & 59.9 & 69.9 & 73.6 & 72.8 \\
    AA-LCR (6\%) & 51.3 & 65.7 & 67.8 & 68.3 \\
    CritPt (6\%) & 0 & 0 & 0 & 0 \\
    AA-Omniscience non-hallucination (4\%) & 86.1 & 85.2 & 86.3 & 85.9 \\
    \midrule
    \multicolumn{5}{@{}l}{\textit{Benchmarks outside the index}} \\
    \addlinespace[3pt]
    IFBench & 61.2 & 70.1 & 70.6 & 71.8 \\
    LiveCodeBench v6 (pass@1) & 50.5 & 66.8 & 71.6 & 71.9 \\
    MATH-500 & 88.8 & 96.6 & 97.2 & 97.2 \\
    \bottomrule
  \end{tabular}
  \endgroup
\end{table}

\FloatBarrier

\subsection{Output Token Accounting and Aggregation}
\label{appendix:koi_output_tokens}

Let $T_b(k)$ be the mean number of output tokens per rollout on benchmark $b$ at loop count $k$, measured with per-loop KV caches. A rollout's count sums completion tokens, including reasoning, over all model requests, agent turns, and sub-steps. It includes requests truncated at the 65,536-token output cap and excludes judges and the $\tau^3$ user simulator. Terminal-Bench counts come from proxy logs, since trajectory records omit truncated requests, and SciCode counts are request-log totals divided by the number of rollouts. GDPval counts include timed-out tasks, which receive no Elo judgment, so its token and score averages cover different task sets. Table~\ref{tab:koi_output_tokens} reports $T_b(k)$.

Figure~\ref{fig:koi_output_tokens} plots the geometric mean of each benchmark's ratio to its one-loop value:
\begin{equation}
  \bar r_k = \left(\prod_{b=1}^{8}\frac{T_b(k)}{T_b(1)}\right)^{1/8}
  \label{eq:koi_output_token_ratio}
\end{equation}
This weights all eight AA-Index benchmarks we run equally, regardless of output length. AA-Omniscience counts once because its two index components share the same generations. Total output tokens over these benchmarks fall by a similar 35.1\% from $k=1$ to $k=4$, and excluding GDPval gives $\bar r_4 = 0.7006$.

\begin{table}[!ht]
  \centering
  \caption{Mean output tokens per rollout for Koi-4B. The last row is $\bar r_k$ (Eq.~\ref{eq:koi_output_token_ratio}).}
  \label{tab:koi_output_tokens}
  \begingroup
  \small
  \setlength{\tabcolsep}{6pt}
  \renewcommand{\arraystretch}{1.15}
  \begin{tabular}{@{}lrrrr@{}}
    \toprule
    Benchmark & $k=1$ & $k=2$ & $k=3$ & $k=4$ \\
    \midrule
    GDPval-AA v2 & 154,205 & 96,979 & 83,979 & 78,506 \\
    Terminal-Bench 2.1 & 215,788 & 146,397 & 121,387 & 107,927 \\
    $\tau^3$-Bench Banking & 35,543 & 27,169 & 24,325 & 22,905 \\
    Humanity's Last Exam & 28,907 & 29,766 & 28,519 & 28,187 \\
    AA-Omniscience & 7,254 & 6,356 & 5,826 & 5,639 \\
    SciCode & 41,061 & 35,408 & 31,861 & 30,253 \\
    GPQA Diamond & 14,059 & 12,604 & 11,835 & 11,502 \\
    AA-LCR & 26,430 & 16,263 & 14,801 & 14,870 \\
    \midrule
    Geometric-mean ratio $\bar r_k$ & 1.0000 & 0.7821 & 0.7046 & 0.6732 \\
    \bottomrule
  \end{tabular}
  \endgroup
\end{table}

Each setting is a single run, and fewer output tokens do not by themselves imply lower inference compute or latency, since each token costs more at higher $k$.

\FloatBarrier

\subsection{Last-Loop KV-Cache Results}
\label{appendix:koi_kv_results}

Table~\ref{tab:koi_kv_cache_intermediate} reports the $k=2,3$ comparisons that complement Table~\ref{tab:koi_kv_cache}. Both cache policies are evaluated on the same checkpoint with the same recipes and a 65,536-token output cap per request, with one run per setting. Each question is repeated 16 times for AA-LCR, 10 for AA-Omniscience, 9 for LiveCodeBench, 8 for GPQA, SciCode, $\tau^3$-Bench Banking, Terminal-Bench, and IFBench, and once for HLE and MATH-500.

\begin{table}[!ht]
  \caption{Per-loop and last-loop KV-cache results at $k=2,3$. Scores are percentages. Output Tokens is the ratio of mean output tokens per rollout, last-loop to per-loop, at the same loop count. Both AA-Omniscience metrics use the same generations.}
  \label{tab:koi_kv_cache_intermediate}
  \centering
  \small
  \setlength{\tabcolsep}{4pt}
  \begin{tabular}{@{}lcccccc@{}}
    \toprule
    \multirow{2}{*}{Benchmark} & \multicolumn{3}{c}{$k=2$} & \multicolumn{3}{c}{$k=3$} \\
    \cmidrule(lr){2-4}\cmidrule(lr){5-7}
      & Per-loop & Last-loop & \shortstack{Output\\[1pt]Tokens} & Per-loop & Last-loop & \shortstack{Output\\[1pt]Tokens} \\
    \midrule
    Terminal-Bench 2.1 & 22.89 & 23.60 & $0.81\times$ & 26.54 & 29.35 & $0.80\times$ \\
    $\tau^3$-Bench Banking & 16.75 & 19.85 & $0.86\times$ & 19.85 & 20.36 & $0.86\times$ \\
    Humanity's Last Exam & 9.08 & 8.94 & $0.95\times$ & 10.10 & 9.87 & $0.94\times$ \\
    AA-Omniscience accuracy & 8.80 & 8.03 & $0.80\times$ & 8.57 & 8.28 & $0.82\times$ \\
    SciCode & 27.95 & 29.12 & $0.99\times$ & 30.82 & 29.99 & $0.94\times$ \\
    GPQA Diamond & 69.95 & 71.91 & $0.97\times$ & 73.61 & 73.55 & $0.96\times$ \\
    AA-LCR & 65.69 & 66.44 & $0.92\times$ & 67.81 & 66.88 & $0.91\times$ \\
    AA-Omniscience non-halluc. & 85.20 & 86.30 & $0.80\times$ & 86.26 & 87.72 & $0.82\times$ \\
    \midrule
    IFBench & 70.12 & 71.06 & $0.90\times$ & 70.65 & 71.46 & $0.93\times$ \\
    LiveCodeBench v6 & 66.79 & 68.75 & $0.94\times$ & 71.59 & 69.21 & $0.97\times$ \\
    MATH-500 & 96.60 & 97.20 & $0.87\times$ & 97.20 & 96.60 & $1.00\times$ \\
    \bottomrule
  \end{tabular}
\end{table}

\paragraph{Score uncertainty.} For the nine benchmarks with per-question scores, each comparison is paired over questions using the per-question mean over repeats. Of the resulting 27 comparisons across $k=2,3,4$, seven reach $|z|\geq2$. Last-loop KV is higher on $\tau^3$-Bench Banking and LiveCodeBench at $k=2$, and lower on LiveCodeBench at $k=3,4$, HLE at $k=4$, and AA-Omniscience accuracy at $k=2,4$. At this uncorrected threshold, about one such result is expected by chance across 27 comparisons. Conversely, $|z|<2$ means no difference is detected, not that the two policies are equivalent. These statistics capture variation across questions within a single run, not across runs.

\paragraph{Output tokens.} Output tokens are counted as in Appendix~\ref{appendix:koi_output_tokens}. Last-loop KV produces fewer output tokens on all ten benchmarks at $k=2$ and on all but MATH-500 at $k=3,4$. Much of the reduction comes from fewer requests truncated at the output cap. At $k=4$, the truncation rate falls from 0.6\% to 0.2\% on Terminal-Bench, 12.5\% to 10.1\% on HLE, 1.3\% to 0.6\% on AA-Omniscience, and 3.3\% to 1.9\% on AA-LCR. On Terminal-Bench, tokens no longer spent on capped requests account for 96--98\% of the reduction, even though mean agent turns rise by 3--7\%.

\paragraph{GDPval-AA v2.} We omit GDPval from Tables~\ref{tab:koi_kv_cache} and~\ref{tab:koi_kv_cache_intermediate}. Tasks that exceed the per-task wall-clock deadline receive no Elo judgment. Last-loop KV fails more of the 220 tasks, mostly by timeout: 13, 29, and 52 failures at $k=2,3,4$, against 1, 6, and 12 with per-loop caches. It also takes 17--19\% more agent turns per task. The two policies' Elo estimates therefore cover different task subsets. The judging reference sets also differ across runs, so per-task rewards cannot be paired. For the same reason, we do not report an AA-Index composite for last-loop KV.

\FloatBarrier

\section{Inference Latency Measurements}
\label{appendix:inference_measurement}

\subsection{Measurement Setup}
\label{appendix:inference_setup}

\begin{figure}[htbp]
  \centering
  \includegraphics[width=0.85\linewidth]{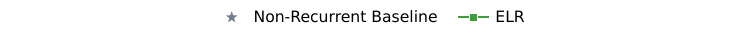}
  \begin{subfigure}[t]{0.4\linewidth}
    \centering
    \includegraphics[width=\linewidth]{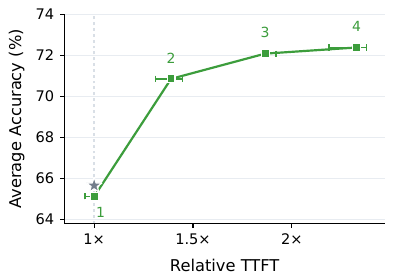}
    \caption{Time to First Token}
    \label{fig:gdp4b_relative_ttft}
  \end{subfigure}
  \hspace{0.1\linewidth}
  \begin{subfigure}[t]{0.4\linewidth}
    \centering
    \includegraphics[width=\linewidth]{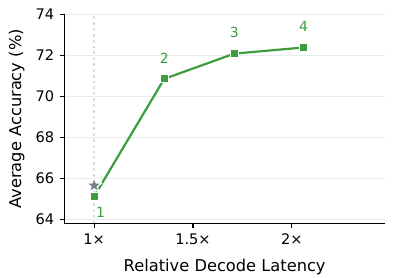}
    \caption{Decode Latency}
    \label{fig:gdp4b_relative_decode}
  \end{subfigure}
  \caption{GDP-Hybrid-4B ten-task average accuracy versus (a) TTFT and (b) per-token decode latency, normalized to the non-recurrent baseline ($1\times$). Numbers indicate loop counts, and horizontal bars show 95\% bootstrap intervals.}
  \label{fig:inference_quality_cost}
\end{figure}

We time the GDP-Hybrid-4B baseline and ELR checkpoints with vLLM 0.20.2 on one H100 80GB GPU, using BF16 weights, eager prefill, and full CUDA-graph decode. The main workload uses batch size one, 1,024 prompt tokens, and 128 greedily generated tokens with EOS ignored, so every request has a fixed output length. TTFT is measured from batch submission to the first output token, and per-token decode latency is the time between the first and last output tokens divided by 127. Inputs are pretokenized, and prefix caching, chunked prefill, speculative decoding, and \texttt{torch.compile} are disabled. We report median times over seven repetitions after two warm-up batches, normalized to the non-recurrent baseline under the same workload. The 95\% confidence intervals in Figure~\ref{fig:inference_quality_cost} are percentile-bootstrap intervals of the time ratio over 10,000 draws, resampling model and baseline repetitions independently. They reflect timing variability only.

\subsection{Sensitivity to Workload}
\label{appendix:inference_sensitivity}

We repeat the measurements at prompt lengths of 128, 1,024, and 4,096 tokens and batch sizes of one and eight, using either CUDA graphs for decode with eager prefill or eager execution throughout. Across prompt lengths and batch sizes, one-loop ELR decode latency differs from the baseline by $-0.26\%$ to $+0.07\%$ with CUDA graphs and by $+0.41\%$ to $+2.16\%$ in eager mode. The corresponding TTFT differences are $-1.54\%$ to $+0.42\%$ with CUDA graphs and $+0.16\%$ to $+1.53\%$ in eager mode.

\section{Route Replay Analysis}
\label{appendix:route_replay}

\subsection{Evaluation Protocol}

We compare the dynamically routed checkpoint (A), replay enabled only during inference on A's checkpoint (B), and a separately trained replay checkpoint (C) from Section~\ref{sec:experiments_route_replay}. Both checkpoints are trained for 300B tokens. We sample 256 validation documents with one 2,048-token window each. Results on 64 separate documents with 8,192-token windows follow the same pattern.

\subsection{Routing and Output Similarity}

Routing statistics compare loops one and four in A (Table~\ref{tab:route_replay_diagnostics}). For each token, the retained fraction is $|S_1\cap S_4|/22$, where $S_1$ and $S_4$ are its first- and fourth-loop sets of 22 selected experts. In contrast, frequency cosine compares aggregate expert-use counts across tokens and does not measure whether a given token keeps its experts. For the layer-output comparison, we hold each layer's fourth-loop input fixed on A's trajectory and compare the output under fresh routes with the output under first-loop routes, without propagating these outputs to later layers. As a control, first-loop routes and weights are circularly shifted within each document, which preserves aggregate usage but breaks token correspondence. Including the shared experts increases output similarity in both comparisons, but replay remains closer to fresh routing than the shifted-route control.

\begin{table}[!ht]
  \centering
  \caption{Cross-loop routing and output comparisons, averaged over ten recurrent MoE layers. Routing statistics compare loops one and four in A, with a 95\% document-bootstrap interval in brackets. Output cosines report first-loop replay / shifted-route control on the same fourth-loop inputs. Full-MoE outputs include shared experts and exclude the outer residual.}
  \label{tab:route_replay_diagnostics}
  \begingroup
  \footnotesize
  \setlength{\tabcolsep}{5pt}
  \renewcommand{\arraystretch}{1.12}
  \begin{tabular}{@{}lc@{}}
    \toprule
    Metric & Value \\
    \midrule
    Retained experts (\%) & 53.17 [52.37, 53.89] \\
    Aggregate frequency cosine & 0.978 \\
    \midrule
    Routed-output cosine & 0.767 / 0.227 \\
    Full-MoE output cosine & 0.918 / 0.718 \\
    \bottomrule
  \end{tabular}
  \endgroup
\end{table}

\subsection{Timing Measurements}

Mixed-depth training update time is averaged over all updates and the reduction from replay stays between 11.0\% and 12.7\% across training intervals. The fixed $k=4$ comparison uses a single checkpoint with replay toggled, with identical code and hardware for both conditions. Inference forward time compares B with A in Megatron-LM on eight H100 GPUs with BF16 weights and expert parallelism of eight. We time full-context forward passes from GPU-resident inputs to logits, taking the maximum wall time across ranks after three warm-up passes. At four loops, replay reduces forward time by 1.21--2.96\% across sequence lengths of 128 to 8,192 tokens and one or two documents per rank, with the smallest reduction at 8,192 tokens.
\section{Ablation Results}
\label{appendix:ablation_results}

This appendix reports the input re-injection ablation on GPT-0.5B, complementing the ablations in Section~\ref{sec:ablations}.

\begingroup
\makeatletter
\setlength{\@fptop}{0pt}
\setlength{\@fpsep}{16pt}
\makeatother

\subsection{Input Re-injection}
\label{appendix:ablation_input_injection}

Table~\ref{tab:ablation_input_injection} compares passing $s_{i-1}$ to $\mathcal{R}_{\theta}$ directly with re-injecting the input-block output as $\tilde{s}_i = s_{i-1} + h^{\mathrm{in}}$ (Section~\ref{sec:method_elr}). All other settings follow the default GPT-0.5B ablation configuration. Repeating the fixed- versus mixed-depth, loop sampling, one-loop override, and delayed mixed-depth ablations with input re-injection gives the same ordering in each (Sections~\ref{sec:ablations_fixed_mixed_training}--\ref{sec:ablations_delayed_mix_depth_training}). We expect the state-update comparison (Section~\ref{sec:ablations_state_update_rule}) to transfer as well.

\begin{table}[h]
  \caption{Input re-injection on GPT-0.5B (validation loss).}
  \label{tab:ablation_input_injection}
  \centering
  \setlength{\tabcolsep}{5pt}
  \begin{tabular}{@{}lccccc@{}}
    \toprule
    Recurrent-block input $\tilde{s}_i$ & $k=1$ & $k=2$ & $k=3$ & $k=4$ & $k=5$ \\
    \midrule
    Non-recurrent baseline & 1.6852 & -- & -- & -- & -- \\
    $s_{i-1}$ & \textbf{1.6830} & \textbf{1.6391} & \textbf{1.6299} & \textbf{1.6278} & \textbf{1.6273} \\
    $s_{i-1} + h^{\mathrm{in}}$ & 1.6858 & 1.6442 & 1.6352 & 1.6329 & 1.6325 \\
    \bottomrule
  \end{tabular}
\end{table}

\FloatBarrier
\clearpage
\endgroup

\end{document}